\RequirePackage{fix-cm}
\documentclass[twocolumn, natbib]{svjour3}          
\smartqed  
\usepackage[ruled]{algorithm2e}
\usepackage{appendix}
\usepackage{array}
\usepackage{float}

\usepackage{amsmath}
\usepackage{amssymb}
\usepackage{booktabs}
\usepackage{graphicx}
\usepackage{lineno}
\usepackage{lipsum}
\usepackage{longtable}
\usepackage{mathtools}
\usepackage[above, below]{placeins}
\usepackage{tabularx}
\usepackage{url}
\usepackage{xcolor}
\usepackage[caption = false]{subfig}
\graphicspath{{SVGs/}}

\newcolumntype{C}{>{\centering\arraybackslash}X}
\newcolumntype{L}{>{\raggedright\arraybackslash}X}
\newcolumntype{R}{>{\raggedleft\arraybackslash}X}

\newcommand{\Diff}{\mathcal{D}}
\newcommand{\dist}{\mathrm{dist}}
\newcommand{\Imm}{\mathcal{I}}
\newcommand{\Shape}{\mathcal{S}}
\newcommand{\srnf}{N}
\newcommand{\Var}{V}
\newcommand{\vol}{A}
\newcommand{\Vol}{\mathrm{Vol}}

\DeclareMathOperator{\Id}{Id}
\DeclareMathOperator{\Tr}{Tr}

\SetKwInOut{KwParam}{Parameters}
\SetKwFunction{DownSample}{DownSample}
\SetKwFunction{GeodesicInterpolation}{GeodesicInterpolation}
\SetKwFunction{InverseSRNF}{InverseSRNF}
\SetKwFunction{Karcher}{Karcher}
\SetKwFunction{LBFGS}{L-BFGS}
\SetKwFunction{Match}{Match}
\SetKwFunction{MultiResolutionMatch}{MultiResolutionMatch}
\SetKwFunction{RandomShuffle}{RandomShuffle}
\SetKwFunction{SRNF}{SRNF}
\SetKwFunction{UpSample}{UpSample}
\SetKwProg{Fn}{function}{}{end}
\SetStartEndCondition{ }{ }{}
\AlgoDontDisplayBlockMarkers
\DontPrintSemicolon
\SetAlCapHSkip{0em}

\journalname{International Journal of Computer Vision}

\newcommand*\patchAmsMathEnvironmentForLineno[1]{%
\expandafter\let\csname old#1\expandafter\endcsname\csname #1\endcsname
\expandafter\let\csname oldend#1\expandafter\endcsname\csname end#1\endcsname
\renewenvironment{#1}%
{\linenomath\csname old#1\endcsname}%
{\csname oldend#1\endcsname\endlinenomath}}
\newcommand*\patchBothAmsMathEnvironmentsForLineno[1]{%
\patchAmsMathEnvironmentForLineno{#1}%
\patchAmsMathEnvironmentForLineno{#1*}}%
\AtBeginDocument{%
\patchBothAmsMathEnvironmentsForLineno{equation}%
\patchBothAmsMathEnvironmentsForLineno{align}%
\patchBothAmsMathEnvironmentsForLineno{flalign}%
\patchBothAmsMathEnvironmentsForLineno{alignat}%
\patchBothAmsMathEnvironmentsForLineno{gather}%
\patchBothAmsMathEnvironmentsForLineno{multline}%
}

\def\makeheadbox{\relax}

\begin{document}

\title{A numerical framework for elastic surface matching, comparison, and interpolation
\thanks{MB is supported by NSF grant 1912037 (collaborative research in connection with 1912030). NC and HH are supported by NSF grants 1819131 and 1945224.}
}

\author{Martin Bauer  \and
        Nicolas Charon \and
        Philipp Harms \and
        Hsi-Wei Hsieh
}

\institute{M. Bauer \at
              Florida State University \\
              Department of Mathematics\\
              \email{bauer@maths.fsu.edu}           
           \and
           N. Charon \at
              Johns Hopkins University \\
              Department of Applied Mathematics and Statistics\\
              \email{charon@cis.jhu.edu}
           \and
           P. Harms \at
              University of Freiburg \\
              Department of Mathematical Stochastics\\
              \email{philipp.harms@stochastik.uni-freiburg.de}   
            \and
           H-W. Hsieh \at
              Johns Hopkins University \\
              Department of Applied Mathematics and Statistics\\
              \email{hhsieh@cis.jhu.edu}     
}

\date{Received: DD Month YEAR / Accepted: DD Month YEAR}

\maketitle

\begin{abstract}
Surface comparison and matching is a challenging problem in computer vision. While elastic Riemannian
metrics provide meaningful shape
distances and point correspondences via the geodesic boundary value problem, solving this problem numerically tends to be difficult. Square root normal fields (SRNF) considerably simplify the computation of certain distances between parametrized surfaces. Yet they leave open the issue of finding optimal reparametrizations, which induce corresponding distances between unparametrized surfaces. This issue has concentrated much effort in recent years and led to the development of several numerical frameworks. In this paper, we take an alternative approach which bypasses the direct estimation of reparametrizations: we relax the geodesic boundary constraint using an auxiliary parametrization-blind varifold fidelity metric. This reformulation has several notable benefits. By avoiding altogether the need for reparametrizations, it provides the flexibility to deal with simplicial meshes of arbitrary topologies and sampling patterns. Moreover, the problem lends itself to a coarse-to-fine multi-resolution implementation, which makes the algorithm scalable to large meshes. Furthermore, this approach extends readily to higher-order feature maps such as square root curvature fields and is also able to include surface textures in the matching problem. We demonstrate these advantages on several examples, synthetic and real.      
\keywords{elastic shape analysis, surfaces, square root normal fields, varifolds.}
\end{abstract}

\tableofcontents

\section{Introduction}
\label{intro}
The analysis of shape-and-image data is an important problem in computer vision. 
Such data arise naturally in many applications such as 
biomedicine \citep{pennec2019riemannian} 
or robotics \citep{turaga2016riemannian}
and is produced e.g.\ by medical imaging devices or time-of-flight cameras in cell phones or cars \citep{grzegorzek2013time}.
While deep-learning techniques have led to significant break-throughs in image analysis, these developments have not yet been paralleled in shape analysis \citep[cf.][]{geirhos2018imagenet}. 
A major difficulty in the direct application of deep-learning methods to shapes is that common shape descriptors are non-unique and vary in structure and dimension. 
For example, one and the same surface can be represented by many different triangular meshes with varying connectivities and numbers of vertices.
While differences in mesh parametrizations matter in numerical computations, they carry no statistical information about the shapes themselves and have to be quotiented out in meaningful statistical analyses. 

Shape analysis provides a mathematical description of shape spaces as quotient spaces in the above sense, as well as a computational toolbox for statistics and machine learning thereon. 
Some classical textbooks are \citet{dryden1998statistical} and \citet{Kendall1999}, and some more recent ones \citet{younes2010shapes},  \citet{srivastava2016functional}, and \citet{pennec2019riemannian}. 
In shape analysis, as in functional data analysis \citep{kokoszka2017introduction}, the starting point is typically a space of functions, say from $S^2$ to $\mathbb R^3$ in the case of surfaces. 

In numerical implementations, this function space is discretized, e.g. by triangular meshes, seen as piecewise linear $\mathbb R^3$-valued functions on some fixed triangulation of $S^2$. 
The shape space corresponding to this function space is determined by the following equivalence relation: 
two functions are considered as representations of the same shape if they differ only by a reparametrization, i.e., by a diffeomorphism on their domain. 
Additionally, rigid motions and scalings are sometimes factored out, as well \citep{Kendall1999}. 
Shape space is a nonlinear infinite-dimensional manifold \citep{cervera1991action},  even if the original function space is linear, and it is the natural configuration space for statistics and machine learning on shapes.

Riemannian geometry is ideally suited as a basis for statistics and machine learning on manifolds of shapes or otherwise:
it allows the estimation of Fr\'echet means and higher-order statistical moments via the geodesic distance function \citep{dryden1998statistical}, 
provides local linearizations and geodesic principal components via the Riemannian logarithm \citep{srivastava2016functional}, 
defines generative random models via stochastic geometric mechanics \citep{pennec2019riemannian}, 
and supplies kernels for support vector machines via the geodesic distance and heat equation \citep{minh2016algorithmic}. 
\phantomsection\label{chg:elastic}

The intuition for the Riemannian setting 
is that two shapes are considered to be similar when they differ only by a small deformation, as measured by a Riemannian metric. 

The Riemannian metrics studied in this paper are often called elastic \citep{younes1998computable, srivastava2011shape, kurtek2012elastic, jermyn2012elastic, jermyn2017elastic} despite the analogy to elastic stretching and bending energies being only loose \citep{rumpf2014geometry, rumpf2015bvariational}.
Alternatively, these metrics are also called inner metrics \citep{bauer2011sobolev} to distinguish them from outer metrics in diffeomorphic shape analysis \citep{younes2010shapes}.
All elastic metrics are invariant to reparametrizations, and this is needed for the metrics to be well-defined on the quotient space of shapes \citep{bauer2016use}. 

Reparametrization-invariance of the metric implies that even linear function spaces have interesting geometries, which are full of surprises.
For example, the geodesic distance of the simplest reparametrization-invariant Riemannian metric vanishes on spaces of curves, surfaces, and diffeomorphisms, as discovered by \citet{michor2005vanishing, michor2006riemannian}.
This degeneracy is a purely infinite-dimensional phenomenon, which led to a systematic study of higher-order Sobolev metrics on mapping spaces. 
It turns out that metrics of order one and higher have non-vanishing geodesic distance \citep{bauer2011sobolev}, 
and metrics of order two and higher are complete on spaces of curves with the same Sobolev regularity \citep{bruveris2014geodesic}.
The metrics considered here are of similar type but lack a zero-order term, which means that they are blind to translations and, beyond dimension one, some further non-trivial deformations \citep{klassen2019closed}.

In data-analytic applications, what is needed foremost are efficient implementations of the geodesic initial and boundary value problems on shape space \phantomsection\label{chg:bvp}(see Section~\ref{sec:elastic_elastic} for definitions and \citet{srivastava2016functional} for applications).
The theoretical understanding of these problems is still incomplete.
The initial value problem is locally well posed for a wide class of Sobolev metrics of order one and higher \citep{bauer2011sobolev}, even of fractional order \citep{bauer2020fractional}.
Moreover, the initial and boundary value problems are globally well-posed for certain second-order Sobolev metrics on spaces of curves \citep{bruveris2015completeness}, but this has not yet been generalized to surfaces. 
Algorithmically, the initial value problem can be solved by a time-discretization of the geodesic equation \citep{bauer2011sobolev} or using discrete geodesic calculus \citep{rumpf2015variational}. 
The boundary value problem on functions (seen as parametrized surface) \phantomsection\label{chg:physics}
is closely related to mesh interpolation \citep{kilian2007geometric,frohlich2011example} and can be solved by the direct method of variational calculus or, for certain elastic metrics, by exploiting isometries to simpler spaces.
For the boundary value problem on shape spaces, one has to solve in addition the optimal correspondence problem, i.e., one has to factor out the action of the  group of reparametrizations. 

Searching for optimal reparametrizations is a challenging task.
Previous approaches relied on a discretization of the reparametrization group and an implementation of the group action on the discretized function space  \citep{srivastava2016functional, jermyn2017elastic,su2020shape}. 
Unfortunately, the implementation of the group action requires the initial and target shapes to be defined on the same pre-specified domain with pre-specified triangulation. 
As this is typically not the case, one first has to solve the parametrization problem \citep{sheffer2007mesh}, 
which is of comparable difficulty to the geodesic boundary problem itself. 
An additional problem is that the reparametrization group can be discretized well for circular domains using monotone correspondences \citep{bernal2016fast} and for spherical domains using spherical harmonics \citep{jermyn2012elastic}, but adaptations to more general domains are difficult. 
Furthermore, compressions of surface patches to points and decompressions of points to surface patches are non-trivial to implement but occur in optimal reparametrizations \citep{lahiri2015precise}. 
Some of these problems can be avoided by minimizing a gauge-invariant energy \citep{tumpach2015gauge,tumpach2016gauge}, but this cannot be used to compute optimal point correspondences, which is one of our main goals.

The main contribution of this paper is a reformulation of the geodesic boundary value problem, which circumvents the above-mentioned problems. 
This reformulation draws on two lines of work: 
square root normal fields and varifold distances. 
Square root normal fields are isometric transformations from spaces of curves or surfaces to simpler manifolds. 
They were introduced by \citet{kurtek2012elastic} and \citet{jermyn2012elastic} and
have their origin in similar transformations for curves, which were discovered by \citet{younes1998computable}, \citet{sundaramoorthi2011new}, and \cite{srivastava2011shape}. 
Square root normal fields map surfaces isometrically into a Hilbert space of square-integrable vector-valued half-densities. 
The Hilbert distance between square root normal fields is a first-order approximation of 
a Riemannian distance and can be computed efficiently by evaluating an integral.
In contrast to general Riemannian
distances, no optimization or time discretization is involved. 
Thus, square root normal fields provide computable (approximate) Riemannian distances with an elastic interpretation.

Varifold distances have their origin in geometric measure theory \citep{federer1969geometric,Almgren1966} and were introduced into the context of diffeomorphic shape analysis by \citet{charon2013varifold}, building on previous work on current distances by \citet{vaillant2005surface} and \citet{glaunes2008large}. 
Their main advantages are that they are fully blind to reparametrizations and can be implemented efficiently on parallel computing architectures, as shown by  \citet{kaltenmark2017general} and \citet{Charon2020fidelity}. 
Despite directly providing distances on shape spaces, the main limitation is that those distances are not the result of a Riemannian metric: in particular, there is no interpretable notion of geodesics between two shapes in this setting.
For this reason, they have to be combined with other shape distances, either on diffeomorphism groups as in diffeomorphic matching \citep{charon2013varifold} or on shape spaces as in elastic matching. 
The latter approach was used for Sobolev metrics on curves by \citet{bauer2017varifold, bauer2018relaxed} and for square root normal distances on curves in the recent conference paper by \citet{bauer2019inexact}, which is a predecessor of the present paper. 
In the present article, the focus is on surfaces, and we add multi-resolution matching, square root normal field inversion, geodesic interpolation, square root curvature fields, surface noise, textured surfaces, non-homotopic surfaces, and Fr\'echet means. Moreover, we provide a stable open-source implementation in \texttt{Python}.

We combine square root normal fields and varifold distances in a new elastic shape matching algorithm, i.e., a new algorithm for solving the geodesic boundary value problem of elastic metrics on shape space. 
The idea is to use varifold distances for relaxing the boundary constraint in the elastic matching problem and to use square root normal fields for boosting the computation of Riemannian distances. 
This has the following advantages. 
\begin{itemize}
\item\label{chg:speed} Speed: Square root normal fields allow one to bypass the time discretization of the geodesic equation, and varifold distances circumvent the costly discretization of reparametrizations. 
\item Applicability and flexibility: The algorithm can be applied directly to simplicial meshes without having to solve the parametrization problem first. It can handle different mesh structures and even different topologies, which is important in the presence of topological noise or for matching shapes with missing parts. Moreover, texture information can be used as in the {\tt fshape} framework of \citet{charon2014functional} and \citet{charlier2017fshape}. 
\item Correctness: The point correspondences found by our algorithm may exhibit compression of surface patches to points and decompression of points to surface patches. Such compressions and decompressions may indeed occur in optimal point correspondences \citep[at least for curves; see][]{lahiri2015precise} and are difficult to model in explicit discretizations of diffeomorphism groups. 
\item Robustness and scalability: The multi-resolution version of our algorithm robustly identifies optimal point correspondences, even for high-resolution surfaces. It scales well to high dimensions thanks to the parallelizability of varifold computations and the linear computational complexity of square root normal fields. 
\end{itemize}
All algorithms are publicly available\footnote{\url{https://github.com/SRNFmatch/SRNFmatch_code}} as a Python package {\tt SRNFmatch}.

\subsection{Structure of the paper}
The paper is structured as follows. 
Section~\ref{sec:method} describes our new problem formulation for the computation of square root normal field distances between surfaces.
Section~\ref{sec:elastic} connects this approach to solving the geodesic boundary value problem of a specific elastic shape metric. 
Section~\ref{sec:implementation} presents the discretization of the algorithm on triangular meshes as well as a multi-resolution scheme to tackle it more efficiently.
Section~\ref{sec:experiments} demonstrates the usefulness of our algorithm on real-world and artificial test data.

\section{Method}
\label{sec:method}

This section describes the proposed new algorithm for the computation of square root normal distances between shapes, using varifold distances as an auxiliary tool. 
An elastic interpretation is given subsequently in Section~\ref{sec:elastic}.
The notation is summarized in Appendix~\ref{sec:notation}.

\subsection{Shape spaces}
\label{sec:shape}

This section defines spaces of parametrized and unparametrized surfaces. 
The latter ones are also called shapes and correspond to equivalence classes of parametrized surfaces modulo reparametrizations. 
Additionally, following \citet{Kendall1999}, one could factor out translations and rotations.
However, as translation and rotation groups are only finite-dimensional, we focus on the more difficult task of quotienting out the infinite-dimensional reparametrization group.

\begin{figure}
\centering
\def\svgwidth{\linewidth}
\begingroup%
  \makeatletter%
  \providecommand\color[2][]{%
    \errmessage{(Inkscape) Color is used for the text in Inkscape, but the package 'color.sty' is not loaded}%
    \renewcommand\color[2][]{}%
  }%
  \providecommand\transparent[1]{%
    \errmessage{(Inkscape) Transparency is used (non-zero) for the text in Inkscape, but the package 'transparent.sty' is not loaded}%
    \renewcommand\transparent[1]{}%
  }%
  \providecommand\rotatebox[2]{#2}%
  \newcommand*\fsize{\dimexpr\f@size pt\relax}%
  \newcommand*\lineheight[1]{\fontsize{\fsize}{#1\fsize}\selectfont}%
  \ifx\svgwidth\undefined%
    \setlength{\unitlength}{626.15193044bp}%
    \ifx\svgscale\undefined%
      \relax%
    \else%
      \setlength{\unitlength}{\unitlength * \real{\svgscale}}%
    \fi%
  \else%
    \setlength{\unitlength}{\svgwidth}%
  \fi%
  \global\let\svgwidth\undefined%
  \global\let\svgscale\undefined%
  \makeatother%
  \begin{picture}(1,0.96028801)%
    \lineheight{1}%
    \setlength\tabcolsep{0pt}%
    \put(0,0){\includegraphics[width=\unitlength,page=1]{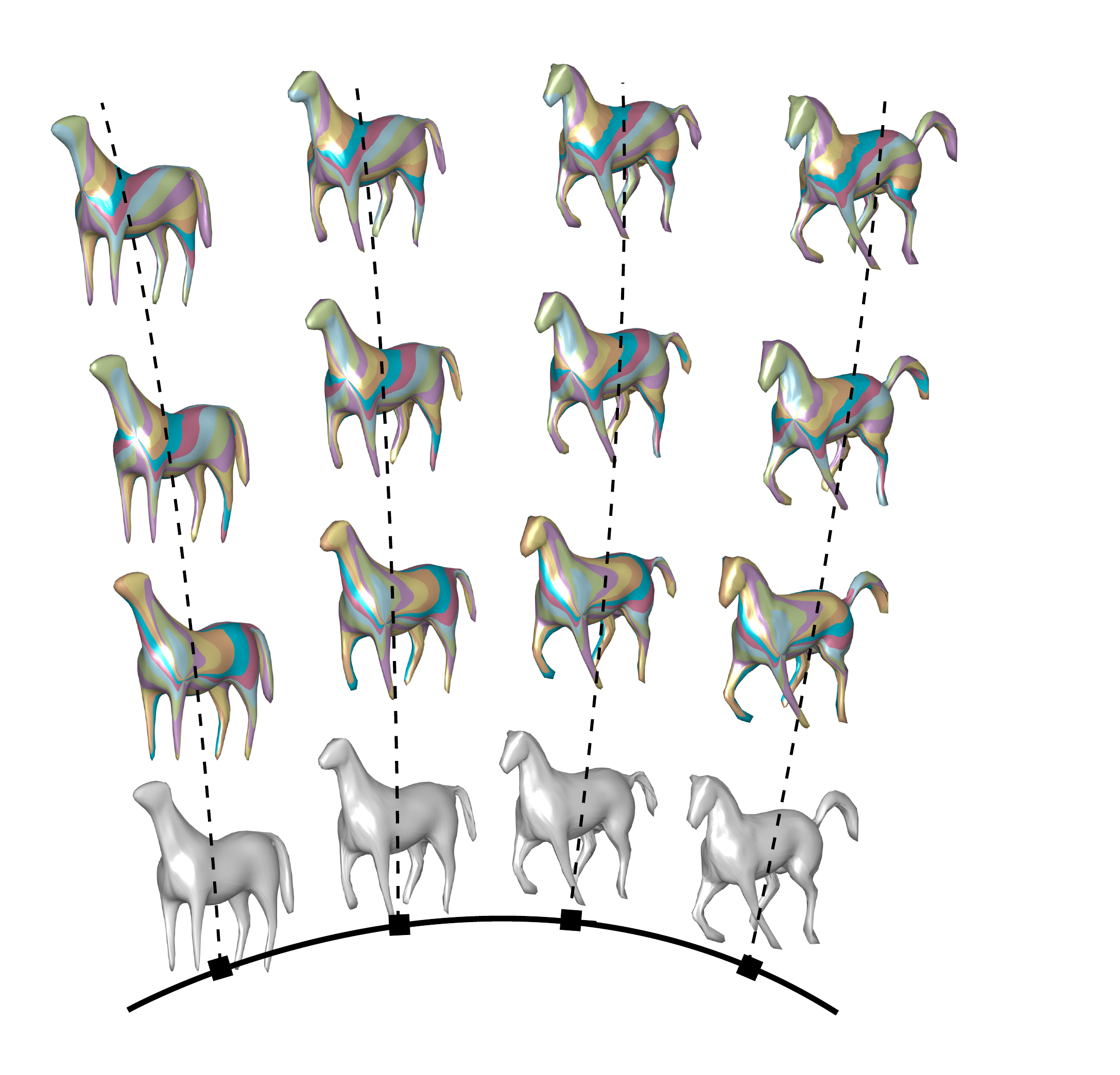}}%
    \put(0.64921207,0.04999611){\color[rgb]{0,0,0}\makebox(0,0)[t]{\lineheight{1.25}\smash{\begin{tabular}[t]{c}$[q]$\end{tabular}}}}%
    \put(0.75724444,0.04572617){\color[rgb]{0,0,0}\makebox(0,0)[lt]{\lineheight{1.25}\smash{\begin{tabular}[t]{l}$\mathcal S$\end{tabular}}}}%
    \put(0.79171115,0.36281136){\color[rgb]{0,0,0}\makebox(0,0)[lt]{\lineheight{1.25}\smash{\begin{tabular}[t]{l}$q\circ\varphi_1$\end{tabular}}}}%
    \put(0.82736489,0.54700641){\color[rgb]{0,0,0}\makebox(0,0)[lt]{\lineheight{1.25}\smash{\begin{tabular}[t]{l}$q\circ\varphi_2$\end{tabular}}}}%
    \put(0.86103643,0.75802627){\color[rgb]{0,0,0}\makebox(0,0)[lt]{\lineheight{1.25}\smash{\begin{tabular}[t]{l}$q\circ\varphi_3$\end{tabular}}}}%
  \end{picture}%
\endgroup%
\caption{Shape space as a quotient of the space of parametrized surfaces. Each shape $[q]$ is an equivalence class of surfaces $q\circ\varphi$, which differ only by a reparametrization $\varphi$. For example, the shape of a horse (gray) is an equivalence class (dashed line) of surfaces with different mesh parametrizations (color-coded). Elastic shape matching requires the computation of the minimal distance between surfaces of one equivalence class to surfaces of another equivalence class. }
\label{fig:shape_space}       
\end{figure}
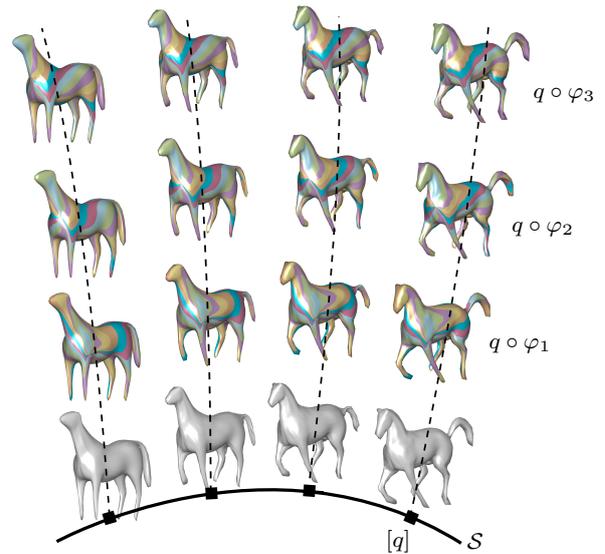

To make this precise, we define the space of parametrized surfaces as the set  $\Imm$ of all oriented immersions $q$ of a $2$-dimensional compact manifold $M$ (possibly with boundary) into $\mathbb R^3$. 
The set $\Imm$ is an open subset of the Fr\'echet space $C^\infty(M,\mathbb R^3)$.
The diffeomorphism group of $M$ is denoted by $\Diff$ and acts on $\Imm$ by reparametrization, i.e., by composition from the right.
The quotient space $\Shape=\Imm/\Diff$, which is called shape space, consists of unparametrized surfaces, i.e., equivalence classes $[q]=\{q\circ\varphi; \varphi \in \Diff\}$.
This is illustrated in Figure~\ref{fig:shape_space}: each equivalence class is a set of parametrized surfaces which correspond to one and the same shape because they differ only by a reparametrizations.
On a discrete level, $\Imm$ corresponds to triangular meshes with fixed connectivity, and reparametrizations correspond to remeshings. 

Unfortunately, the quotient space $\Shape$ has no straight-forward discretization, and this constitutes one of the main difficulties in numerical shape analysis.

\subsection{Square root normal fields}
\label{sec:srnf}
Square root normal fields are part of a family of feature maps, which can be used to define shape distances related to elastic metrics, as we explain in Section~\ref{sec:srnf_metric}.
The first instance of such a feature map, which was discovered by \citet{younes2008metric} and \citet{sundaramoorthi2011new}, is the map from planar curves $q$ to the complex square root $\sqrt{q'}$ of their velocity, seen as an element of an infinite-dimensional Stiefel manifold.
Similarly, \citet{srivastava2011shape} considered the map from planar curves $q$ to their square root velocity field $q'|q'|^{-1/2}$, seen as an element of a Hilbert space. 
Subsequently, \citet{kurtek2012elastic} and \citet{jermyn2012elastic} found an extension to higher dimension, namely the map from oriented hyper-surfaces $q$ to their square root normal field (SRNF) $\srnf_q$,  which is the unit normal field $n_q$ multiplied by the square root of the Riemannian area form $\vol_q$:  
\begin{equation*}
\srnf_q \coloneqq n_q\; \vol_q^{1/2}\;. 
\end{equation*}
For curves, the SRNF coincides up to a rotation of 90 degrees with the square root velocity field. 
For two-dimensional surfaces, it is given in coordinates $(u,v) \in \mathbb R^2$ as
\begin{equation}
\label{equ:srnf_coords}
\srnf_q=(q_u\times q_v)\; |q_u\times q_v|^{-1/2}\;, 
\end{equation}
where the subscripts denote partial derivatives,  $\times$ denotes the cross product on $\mathbb R^3$, and $|\cdot|$ denotes the norm on $\mathbb R^3$. 
Geometrically, the SRNF belongs to the space of square-integrable $\mathbb R^3$-valued half-densities, which is a Hilbert space with the $L^2$ scalar product \citep[see][Section~10.4 for further details]{Michor08}. 
Half-densities can be identified with real-valued functions by fixing a reference half-density.
This shall be done implicitly in all numerical discretizations. 
However, it should be kept in mind that half-densities transform differently than functions under reparametrizations: for any $\varphi \in \Diff$, in any coordinate system on $M$, 
\begin{equation}
\label{equ:half-density-transformation}
\srnf_{q\circ\varphi}
=
(\srnf_q)\circ\varphi \ \det(D\varphi)^{1/2}\;,
\end{equation}
where $D\varphi$ denotes the differential (or Jacobian) of the diffeomorphism $\varphi$.

The SRNF distance between parametrized surfaces $q_0$ and $q_1$ is defined as the $L^2$ distance between the associated SRNFs:
\begin{equation}
\label{equ:srnf_dist}
\dist_N(q_0,q_1)
\coloneqq
\int_M |\srnf_{q_0}-\srnf_{q_1}|^2\;.
\end{equation}
Note that the right-hand side is well-defined because the square of a half-density is a density. 
The SRNF distance is sometimes called chordal distance to point out that the SRNF map is non-surjective and that consequently the SRNF distance differs from the intrinsic distance on its range. 
The SRNF distance has two advantages: it can be computed easily, as described in Section~\ref{sec:discretization}, and it is $\Diff$-invariant, i.e., for any $\varphi \in \Diff$, 
\begin{equation}
\label{equ:srnf_diff_invariance}
\dist_N(q_0\circ\varphi,q_1\circ\varphi)
= 
\dist_N(q_0,q_1)\;.
\end{equation} 
This follows from \eqref{equ:half-density-transformation} and the change-of-variable formula for integrals.
However, this distance is not yet a distance between shapes, and it remains to quotient out reparametrizations.
Accordingly, the SRNF distance between shapes $[q_0],[q_1]\in\Shape$ is defined as the minimal SRNF distance between all reparametrizations of $q_0$ and $q_1$, i.e.,
\begin{equation}
\label{equ:dist_srnf_shape}
\begin{aligned}
\dist_N([q_0],[q_1])
&\coloneqq
\inf_{\varphi_0,\varphi_1\in \Diff} \dist_N(q_0\circ\varphi_0,q_1\circ\varphi_1)
\\&=
\inf_{\varphi\in \Diff} \dist_N(q_0,q_1\circ\varphi)\;,
\end{aligned}
\end{equation}
where the second equation follows from the $\Diff$-invariance \eqref{equ:srnf_diff_invariance}. 
Similar distances can be defined for more general $\Diff$-equivariant feature maps, as discussed in Section~\ref{sec:generalized}.
An important caveat is that the SRNF distance is only a pseudo-distance because the SRNF map is non-injective, even after factoring out translations; see \citet{klassen2019closed}.
We will not dwell upon this issue and freely speak of distances even when they are only pseudo-distances. 
The main difficulty in \eqref{equ:dist_srnf_shape} is the search for an optimal reparametrization $\varphi\in\Diff$.  
We address this problem using varifold distances, which are described next.

\subsection{Varifold distances}
\label{sec:varifold}
Geometric measure theory provides several embeddings of shape spaces into some Banach spaces of distributions, including currents \citep{vaillant2005surface, glaunes2008large}, varifolds \citep{charon2013varifold}, and normal cycles \citep{roussillon2016kernel}, see \citep{Charon2020fidelity} for a recent review. In this paper, we adopt the framework of \citet{kaltenmark2017general}.
The varifold $\mu_q$ associated to $q \in \Imm$ is the image measure $(q,n_q)_*\vol_q$ on $\mathbb R^3\times S^{2}$, where $n_q$ is the unit normal field of $q$, and $\vol_q$ is the area form (also known as surface measure) of $q$. 
In other words, for any Borel set $B \subset \mathbb R^3\times S^{2}$, $\mu_q(B)$ is the area of all $m \in M$ such that $(q(m),n_q(m))$ belongs to $B$. 
Importantly, this measure representation does not depend on the parametrization of $q$ and thus provides an embedding of shape space $\Shape$ into the space of positive measures of $\mathbb R^3\times S^{2}$. 
Then, given a norm $\|\cdot\|$ on the space of measures, one defines the varifold distance
between $[q_0],[q_1] \in \Shape$ as
\begin{equation}
\label{equ:dist_var}
\dist_\Var(q_0,q_1)
\coloneqq
\|\mu_{q_0}-\mu_{q_1}\|\;.
\end{equation}
While there are many possible distances that one can introduce on spaces of measures, norms defined from positive definite kernels on $\mathbb R^3\times S^{2}$ have been shown to lead to particularly advantageous expressions for numerical computations. 
Specifically, following the setting of \citet{kaltenmark2017general}, we consider the class of norms $\|\cdot\|_{V^*}$, where $V$ is a reproducing kernel Hilbert space of functions on $\mathbb R^3\times S^{2}$, whose kernel is of the form
\begin{equation}
\label{equ:kernel_var}
k(x_1,n_1,x_2,n_2) \coloneqq \rho(|x_1-x_2|) \gamma(n_1 \cdot n_2)\;, 
\end{equation}
where $\rho$ and $\gamma$ are two functions defining a radial kernel on $\mathbb{R}^3$ and a zonal kernel on $S^{2}$, respectively.  
Then the scalar product between varifolds $\mu_{q_0}$ and $\mu_{q_1}$ can be shown to be:
\begin{multline}
\label{equ:norm_var}
\langle \mu_{q_0},\mu_{q_1}\rangle_{V^*}
=
\iint_{M \times M} \rho\big(|q_0(u_0,v_0)- q_1(u_1,v_1)|\big) 
\\
\gamma\big(n_{q_0}(u_0,v_0)\cdot n_{q_1}(u_1,v_1)\big) \vol_{q_0}(u_0,v_0) \vol_{q_1}(u_1,v_1)
\end{multline}
Accordingly, the varifold distance between any two triangular surfaces $q_0$ and $q_1$ can be computed as 
\begin{equation*}
\dist_\Var(q_0,q_1)^2 
=
\|\mu_{q_0}\|_{V^*}^2-2\langle \mu_{q_0},\mu_{q_1} \rangle_{V^*} + \|\mu_{q_1}\|_{V^*}^2 \;.
\end{equation*}

As shall be see in Section~\ref{sec:implementation}, the previous expression has a natural and simple discrete equivalent for triangular meshes. 

Varifold distances are equivariant to the action of rigid motions: for any $R\in SO(3)$ and $h\in \mathbb{R}^3$, 
\begin{equation*}
\dist_\Var(R q_0 + h,Rq_1+h) = \dist_\Var(q_0,q_1)\;. 
\end{equation*} 
More importantly, they are $(\Diff\times\Diff)$-invariant: for any $\varphi_0,\varphi_1\in\Diff$, 
\begin{equation*}
\dist_\Var(q_0\circ\varphi_0,q_1\circ\varphi_1)
=
\dist_\Var(q_0,q_1)\;.
\end{equation*}
There is an important difference between the $\Diff$-invar\-iance of 
the SRNF distance and the $(\Diff\times\Diff)$-invariance of varifold distances: 
a $\Diff$-invariant distance vanishes at $q_0,q_1\in\Imm$ if and only if $q_0=q_1$, whereas a $(\Diff\times\Diff)$-invariant distance vanishes at $q_0,q_1\in\Imm$ if and only if $q_0=q_1\circ\varphi$ for some $\varphi\in\Diff$.
Varifold distance are rather ideally suited for enforcing the latter condition, i.e., membership of the same $\Diff$-orbit in $\Imm$.
This makes them highly useful in shape analysis, as described next.
However, they are not tied to a Riemannian metric and thus are not associated to geodesics between shapes.

Another interesting feature of the varifold framework is the possibility to incorporate additional texture information in the metric, as detailed in \citep{charlier2017fshape}. Assume that the given immersed shape $q$ carries a scalar texture map $\zeta\colon M \rightarrow \mathbb R$. Then the couple $(q,\zeta)$ can be represented as a functional varifold, i.e., as the image measure 
\begin{equation*}
\mu_{q,\zeta} 
\coloneqq
(q,n_q,\zeta)_*\vol_q 
\end{equation*}
on $\mathbb R^3\times S^{2} \times \mathbb{R}$. The previous kernel metrics can then be readily extended to this new situation using the following new kernel:
\begin{equation*}
k(x_1,n_1,\zeta_1,x_2,n_2,\zeta_2) \coloneqq 
\rho(|x_1-x_2|) \gamma(n_1 \cdot n_2) \tau(|\zeta_1 - \zeta_2|)\end{equation*}
Here, $\tau$ defines a positive-definite kernel on $\mathbb{R}$, which adds to \eqref{equ:kernel_var} a notion of proximity between texture values. 
The corresponding functional varifold distance  is defined similarly to \eqref{equ:norm_var} and \eqref{equ:dist_var}. 
It provides a distance between textured surfaces $(q_0,\zeta_0)$ and $(q_1,\zeta_1)$, which can be used to enforce the joint constraint of matching geometries $q_0=q_1\circ\varphi$ and matching textures $\zeta_0 = \zeta_1 \circ \varphi$, up to reparametrization by some $\varphi \in \Diff$.  

\subsection{Combining SRNF and varifold distances}
\label{sec:elastic-varifold}

From a numerical perspective, the main difficulty in the computation \eqref{equ:dist_srnf_shape} of SRNF distances between shapes is the search for an optimal reparametrization $\varphi\in\Diff$ in the diffeomorphism group of $M$. 
For curves, reparametrizations of $M=S^1$ are monotone correspondences, and the optimal correspondence can be found using dynamic programming, following \citet{frenkel2003curve} and \citet{sebastian2003aligning}.
For surfaces, diffeomorphisms of $M=S^2$ can be represented using spherical harmonics, and the optimal harmonic representation can be found via gradient descent \citep{srivastava2011shape,kurtek2012elastic,jermyn2012elastic}. 
To implement the action of diffeomorphisms on shapes, these algorithms presuppose that the shapes are already para\-metrized consistently, i.e., defined on the same mesh. 
Otherwise, one has to solve the challenging parametrization problem first, which is of comparable difficulty to the search for an optimal reparametrization. 
An additional disadvantage is that these methods do not generalize easily to domains other than $S^1$ or $S^2$. 

We circumvent these problems by exploiting the $(\Diff\times\Diff)$-invariance of varifold distances. 
Previously, \citet{charon2013varifold} and \citet{kaltenmark2017general} applied a similar strategy to diffeomorphic matching. 
Moreover, \citet{bauer2018relaxed} used varifold distances for the first time in the context of elastic matching, albeit only for curves.
The starting point is the observation that problem~\eqref{equ:dist_srnf_shape} is equivalent to the constrained optimization problem
\begin{equation*}
\min_{\tilde q_1\in\Imm} 
\hspace{1ex}
\dist_N(q_0,\tilde q_1)
\quad
\text{subject to}
\quad
\dist_{\Var}(\tilde q_1,q_1)=0\;.
\end{equation*}
The optimization over $\varphi\in\Diff$ in \eqref{equ:dist_srnf_shape} has now been replaced by an optimization over $\tilde q_1\in\Imm$ subject to a varifold constraint, which ensures that $\tilde q_1$ belongs to the $\Diff$-orbit of $q_1$, as required.
Relaxation of the constraint using a (large) penalty parameter $\lambda$ yields
\begin{equation}
\label{equ:elastic_var}
\min_{\tilde q_1\in\Imm} 
\hspace{1ex}
\dist_N(q_0,\tilde q_1) + \lambda \dist_{\Var}(\tilde q_1,q_1)\;.
\end{equation}
At this point one may wonder whether both of the two distances in \eqref{equ:elastic_var}, namely the varifold and SRNF distance, are really needed. 
This is indeed the case, as the formulation exploits---and requires---both the $(\Diff\times\Diff)$-invariance of the varifold distance and the $\Diff$-invariance of the SRNF distance.
In other words, the SRNF
distance alone cannot explore different parametrizations, 
while the varifold distance alone cannot distinguish between good and bad parametrizations and does not define geodesics between shapes. 

\subsection{Symmetric elastic-varifold matching}

In the category of smooth surfaces and diffeomorphisms, the asymmetric matching problem \eqref{equ:elastic_var} is equivalent to the following symmetrized matching problem:
\begin{multline}
\label{equ:var_elastic_var}
\min_{\tilde q_0,\tilde q_1\in\Imm} 
\hspace{1ex}
\lambda \dist_{\Var}(q_0,\tilde q_0)
+\dist_N(\tilde q_0,\tilde q_1) 
\\
+ \lambda \dist_{\Var}(\tilde q_1,q_1)\;.
\end{multline}
However, for non-smooth surfaces and diffeomorphisms the two problems are different, and the symmetric version offers significant advantages. 
The symmetric matching problem  \eqref{equ:var_elastic_var} is more general, as it is well-defined for arbitrary varifolds $q_0$ and $q_1$.
As an aside, it should be kept in mind that the matching energy \eqref{equ:var_elastic_var} depends on the choice of $M$. 
This choice can be interpreted as a topological prior and corresponds on a discrete level to a choice of triangular mesh for the surfaces $\tilde q_0$ and $\tilde q_1$. 
A further advantage is that the symmetric matching problem allows for reparametrizations of both surfaces $q_0$ and $q_1$. 
In the case of curves this flexibility is needed to represent optimal reparametrizations, which may compress a patch to a point or conversely decompress a point to a patch \citep{lahiri2015precise}. 
Similar compressions and decompressions are to be expected in the case of surfaces and cannot be modeled in the asymmetric matching problem, where only one of the surfaces is reparametrized. 

\subsection{Comparison to alternative methods}
\label{sec:comparison}

Compared to alternative methods of surface comparison and matching, program \eqref{equ:var_elastic_var} has the following advantages:
\begin{itemize}
\item The reparametrization group $\Diff$ does not need to be discretized, and its action on $\Imm$ does not need to be implemented. This allows one to work with simplicial meshes without having to solve the parametrization problem first, as for instance in \citet{srivastava2016functional} and \citet{jermyn2017elastic}. Moreover, it easily generalizes to domains $M$ more general than $S^1$ or $S^2$, which is not without difficulties when dynamic programming or spherical harmonic are used, as in  e.g.\ \citet{bernal2016fast} or \citet{jermyn2012elastic}. 

\item The varifold matching term provides the flexibility for handling topological noise, matching shapes with missing parts, and incorporating texture information, similarly to the {\tt fshape} framework \citep{charon2014functional,charlier2017fshape}. 

\item The computations are faster than for general Riemannian
distances (see Section~\ref{sec:elastic}), where an additional time discretization is needed. 
\end{itemize}
These advantages are also demonstrated by the numerical examples in Section~\ref{sec:experiments}.

\subsection{Applications}

Program \eqref{equ:var_elastic_var} is the basis for several important tasks in shape analysis:
\begin{itemize}
\item It implements the geodesic boundary value problem for an elastic shape metric, which is described in further detail in Section~\ref{sec:elastic}.
At intermediate time points 
the geodesic can be obtained from the linear interpolation of the SRNFs
by approximate inversion of the SRNF map $q\mapsto\srnf_q$, as described in  Section~\ref{sec:srnf_inversion}.

\item The geodesic boundary value problem is the basis for machine learning and statistics on manifolds. It provides geodesic distances, which are used for agglomerative clustering, support vector machines, Fr\'echet means, and higher-order moments; cf.~\cite{srivastava2016functional}. Moreover, it implements the Riemannian logarithm, which is used in geodesic principal component analysis and statistics on manifolds; cf.~\citet{bhattacharya2012nonparametric} and \citet{pennec2019riemannian}. The Riemannian logarithm at the reparametrized surface $\tilde q_0$ is the initial velocity of the above-mentioned horizontal geodesic. Moreover, the Riemannian logarithm at the original surface $q_0$ can be obtained from the asymmetric version \eqref{equ:elastic_var} of the program.  

\item The program implements the matching problem, i.e., the optimizers $\smash{\tilde q_0}$ and $\smash{\tilde q_1}$ describe a point correspondence between the given shapes $q_0$ and $q_1$. 

\item The program implements the parametrization problem, i.e., the optimizers $\tilde q_0$ and $\tilde q_1$ parametrize the given shapes $q_0$ and $q_1$ as piece-wise linear functions on a given mesh. As an aside, this parametrization is not determined uniquely by \eqref{equ:var_elastic_var} and could therefore be selected freely using some additional criteria of angle, distance, or area preservation, which have been developed in the literature on mesh parametrization; see e.g.\ \citet{floater2005surface} or \citet{sheffer2007mesh}. 
\end{itemize}

\section{Elastic interpretation}
\label{sec:elastic}

The surface matching approach presented in the previous section approximately solves the geodesic boundary value problem of a certain first-order elastic metric on the shape space of surfaces, as we explain next.

\subsection{Elastic shape analysis}
\label{sec:elastic_elastic}
To emphasize this connection, we start with some background on elastic metrics, referring to the surveys by \citet{bauer2014overview} or \citet{jermyn2017elastic} for further details and references. 
Elastic shape analysis operates in a Riemannian framework, where infinitesimal shape deformations are measured by a Riemannian metric, which is often related to an elastic (or plastic) deformation energy. 
Given a Riemannian metric $G$ on $\Imm$, one defines the Riemannian distance between two shapes $q_0,q_1 \in \Imm$ as 
\begin{equation}
\label{equ:dist_imm}
\dist_G(q_0,q_1)^2
\coloneqq
\inf_{\substack{q\in C^\infty([0,1],\Imm)\\q(0)=q_0, q(1)=q_1}}\ \int_0^1 G_q(\partial_t q,\partial_t q) dt\;. 
\end{equation}
The optimizers in \eqref{equ:dist_imm} are constant-speed geodesics in $\Imm$. 
If $G$ is $\Diff$-invariant, as shall be assumed throughout, one obtains also a distance between shapes $[q_0],[q_1]\in \Shape$ by setting
\begin{equation}
\label{equ:dist_bi}
\begin{aligned}
\dist_G([q_0],[q_1])
\coloneqq&
\inf_{\varphi_0,\varphi_1\in \Diff} \dist_G(q_0\circ\varphi_0,q_1\circ\varphi_1)
\\=&
\inf_{\varphi\in \Diff} \dist_G(q_0,q_1\circ\varphi)\;,
\end{aligned}
\end{equation}
where the second equation follows from the $\Diff$-invariance of the Riemannian metric $G$.
This is the same quotient construction as for SRNF distances \eqref{equ:dist_srnf_shape}, and the optimization over reparametrizations can again be avoided using varifold distances similarly to \eqref{equ:var_elastic_var}. 
However, in contrast to SRNF distances and elastic deformation energies in physics \citep{rumpf2014geometry, rumpf2015bvariational}, the Riemannian 
distance \eqref{equ:dist_bi} retains a dynamical interpretation, as its optimizers are constant-speed geodesics in $\Shape$. 

For low-order Sobolev metrics $G$, the distances on $\Imm$ and $\Shape$ do not separate points, as discovered by \citet{michor2007overview}. 
Fortunately, this degeneracy disappears for metrics of order one and higher \citep{bauer2011sobolev}.
For simplicity, we will speak of Riemannian metrics and their associated distances even when they are only pseudo-metrics and pseudo-dis\-tan\-ces. 

From an applied perspective, efficient numerical implementations of the optimization problems \eqref{equ:dist_imm} and \eqref{equ:dist_bi} are crucial, as they are the algorithmic basis for the computation of point correspondences, Fr\'echet means, geodesic principal components, parallel transport, and so on; see  e.g.~the survey of \citet{bauer2014overview} or the book of \citet{srivastava2016functional}.
For general elastic metrics, problem \eqref{equ:dist_imm} can be implemented using path straightening methods. 
This involves a time discretization and is computationally costly.
However, for certain elastic metrics, which are related to SRNFs, this time discretization can be avoided, as described next. 

\subsection{Square root normal metrics}
\label{sec:srnf_metric}

Recall that the SRNF is a map from $q \in \Imm$ to the Hilbert space of square-integrable $\mathbb R^3$-valued half-densities. 
The SRNF metric on $\Imm$ is the pull-back of the Hilbert scalar product along this map. 
Thus, it is defined for any tangent vector $h\in T_q\Imm$ as 
\begin{equation}
\label{equ:srnf_metric}
G_q(h,h) 
\coloneqq
\int_M | D_{(q,h)}\srnf_q|^2\;,
\end{equation}
where $D_{(q,h)}\srnf_q$ denotes the directional derivative of $\srnf_q$ at $q$ in the direction $h$.
Note that the right-hand side in \eqref{equ:srnf_metric} is well-defined since the square of a half-density is a density.

The SRNF metric belongs to the class of first order Sobolev metrics, which have been studied in great detail by \citet{michor2007overview}, \citet{mennucci2008properties}, \citet{bauer2011sobolev}, and many others.
An explicit formula for it is established in Appendix~\ref{sec:srnf_formula}:
\begin{equation}
\label{equ:srnf_metric_formula}
G_q(h,h) 
=
\int_M \Big(\big|(\nabla h)^\bot\big|^2 + \frac14 \Tr\big((\nabla h)^\top\big)^2 \Big)\vol_q\;.
\end{equation}
Here $\nabla$ is the coordinate-wise derivative of $\mathbb R^3$-valued functions, $\bot$ and $\top$ are the normal and tangential projections satisfying $\bot+Tq\circ\top=\Id_{\mathbb R^3}$, the norm $|\cdot|$ is computed with respect to the pull-back cometric, and $\Tr$ denotes a trace; see Appendix~\ref{sec:notation} for further notation.

While computing geodesic distances of general elastic metrics can be quite cumbersome, this is much simpler for SRNF metrics thanks to the approximation 
\begin{equation}
\label{equ:approx}
\dist_G(q_0,q_1)
\approx
\dist_N(q_0,q_1)\;,
\end{equation}  
where $\dist_G$ is the Riemannian distance \eqref{equ:dist_imm} of the SRNF metric \eqref{equ:srnf_metric}, and $\dist_N$ is the SRNF distance \eqref{equ:srnf_dist}.
The approximation is exact whenever the straight line between $\srnf_{q_0}$ and $\srnf_{q_1}$ is contained in the range of the SRNF map $q\mapsto\srnf_q$.
The reason is that the SRNF map is a Riemannian isometry, by construction, and that geodesics in the Hilbert space of SRNFs are straight lines.
In general, the approximation is exact up to first order for $q_0$ close to $q_1$; see Appendix~\ref{sec:approximation} for a proof.
 Some generalizations to higher-order metrics are discussed next.

\subsection{Generalized square root normal fields}
\label{sec:generalized}

The theoretical and algorithmic framework of the previous sections can be extended using more general feature maps than SRNFs.
For example, \citet{jermyn2012elastic} introduced the Gauss map $q\mapsto(g_q,n_q)$ as a feature map, where $g_q$ is the pull-back metric of $q$. 
This feature map is injective up to isometries of $\mathbb R^3$ by a classical result of \citet{abe1975isometric}, and the geodesic distance on the image space has recently been computed by \citet{su2020simplifying}.
Alternatively, \citet{bauer2011sobolev} used the map $q \mapsto dq$ as a feature map, where $dq$ is the derivative of $q$, seen as an $\mathbb R^3$-valued one-form on $M$. 
For simplicity we restrict ourselves to feature maps which transform under reparametrizations similarly to half-densities. 
As explained below, this automatically guarantees $\Diff$-invariance of the $L^2$ distance between features, which is particularly easy to compute. 

The general setup is as follows.
One considers $\Diff$-equivariant maps $\Phi$ from $\Imm$ to the Hilbert space of square-integrable sections of $E\otimes\Vol^{1/2}$, where $E$ is a tensor bundle over $M$ with $\Diff$-invariant fiber metric $\langle \cdot,\cdot\rangle_E$ and corresponding fiber norm $|\cdot|_E$, and where $\Vol^{1/2}$ is the half-density bundle over $M$. 
The corresponding distance, which generalizes the SRNF distance \eqref{equ:srnf_dist}, is defined as
\begin{equation}
\label{equ:feature_dist}
\dist_\Phi(q_0,q_1)^2
\coloneqq
\int_M |\Phi_{q_0}-\Phi_{q_1}|_E^2\;.
\end{equation}
Note that the integral on the right-hand side is well-defined because the square of a half-density is a density. 
Importantly, the distance can be computed easily from fiber-wise distances of the bundle $E$ thanks the $L^2$ structure \citep[cf.][]{Ebin1970b}.
In particular, if $\langle\cdot,\cdot\rangle_E$ is flat on each fiber of $E$, then the computation reduces to the evaluation of an integral. 
Moreover, thanks to the assumption that $\Phi$ transforms as a half-density, the distance \eqref{equ:feature_dist} is $\Diff$-invariant, i.e., for all $\varphi\in\Diff$,
\begin{equation*}
\dist_\Phi(q_0\circ\varphi,q_1\circ\varphi)
=
\dist_\Phi(q_0,q_1)\;.
\end{equation*} 
The corresponding $\Diff$-invariant elastic metric, which generalizes the SRNF metric \eqref{equ:srnf_metric}, is given by
\begin{equation}
\label{equ:feature_metric}
G_q(h,h)
\coloneqq
\int_M|D_{(q,h)}\Phi_q|_E^2\;,
\end{equation}
where $h \in T_q\Imm$ is a tangent vector to $q$.
An example is provided next.

\subsection{Example: square root curvature fields}
\label{sec:SRCF}
 
We introduce a new second-order feature map, which we call square root curvature field (SRCF).
Combined with the SRNF feature map, it yields a meaningful second-order elastic metric, whose geodesic distance can be computed efficiently similarly to \eqref{equ:approx}.
The SRCF is defined as 
\begin{equation}
\label{equ:srcf_vector}
\Phi_q\coloneqq H_q\;\vol_q^{1/2}, 
\end{equation}
where $H_q$ is the vector-valued mean curvature of $q$. 
This feature map $\Phi$ simultaneously encodes information about the normal vector, the area form, and the curvature of the shape. 
The SRCF distance \eqref{equ:feature_dist} 
coincides with the \citet{willmore1993riemannian} energy of $q_1$ if $q_0$ is a minimal surface. 
More generally, this distance quantifies differences in the mean curvatures of $q_0$ and $q_1$, weighted symmetrically using the respective area forms of $q_0$ and $q_1$. 
The associated SRCF metric \eqref{equ:feature_metric} is a second-order elastic metric, whose highest-order term is the normal component of the Laplacian $\Delta_q$ on the surface $q$:
\begin{equation*}
G_q(h,h)
=
\int_M |(\Delta_q h)^\bot + C(\nabla h)|^2 \vol_q\;,
\end{equation*}
where $C(\nabla h)$ denotes a first-order term, which is a contraction of $\nabla h$ with the metric, cometric, and second fundamental form; see Appendix~\ref{sec:srcf_metric}.
The same form of metric is obtained using the scalar instead of the vector-valued mean curvature, but the feature map contains less information in this case. 
Despite the complicated form of the SRCF metric, its geodesic distance is easy to compute, thanks to the Hilbert structure on the range of the feature map. 
We also experimented with alternative forms of the SRCF, where the vector-valued mean curvature is replaced by the scalar mean curvature; see Section~\ref{sec:discretization}.

\section{Implementation}
\label{sec:implementation}

\label{sec:discretization}
This section describes the implementation of the geodesic boundary value problem, i.e., the surface matching algorithm for triangular meshes. 
The term geodesic shall be used quite liberally, even when approximations, relaxations, and discretizations of the Riemannian energy functional are involved.

\subsection{Open source library}
All algorithms described in this section are publicly available\footnote{\url{https://github.com/SRNFmatch/SRNFmatch_code}} as a Python package {\tt SRNFmatch}. 
These include the routines for symmetric and asymmetric surface matching based on SRNF and SRCF energies, inverting the SRNF and SRCF maps, computing geodesic interpolations, and generating some of the figures in this paper.
Our implementation relies on the following Python libraries: {\tt Numpy}, {\tt Scipy}, {\tt Pytorch}, {\tt PyKeops}, and {\tt Vtk}.
We refer to the online documentation for further details on the code.   

\subsection{Mesh discretization}
\label{sec:mesh}

To discretize the space of surfaces, we use a triangular surface as the domain $M$ of the function space $\Imm$ and restrict this function space to piece-wise linear functions. 
Alternative disretizations for more general mesh structures could be used, as well.
Any such mesh structure defines a natural reference density on $M$, which assigns weight $1$ to every face of the mesh. 
This reference density and its corresponding half-density are used throughout to identify densities and half-densities with functions.
Alternative reference densities would lead to the same results but involve additional quadrature weights.

In the computation of SRNF or SRCF distances, the triangulation of $M$ can be chosen freely but has to coincide for the initial and target surfaces.
In contrast, in the geodesic boundary value problem, the initial and final surfaces may have different mesh structures and even different topologies. 
This flexibility is granted by the varifold terms.

\subsection{SRNF distances between triangular meshes}
The SRNF of a triangular mesh is the piece-wise constant function \eqref{equ:srnf_coords}, which associates to any face of the mesh the square root of its area multiplied by the unit-normal vector. 
Thus, the SRNF of a surface $q \in \Imm$ can be written as a function $\srnf_{q,f}$ of the faces $f$ of the mesh: 
\begin{align*}
\srnf_{q,f} 
\coloneqq 
n_{q,f}\; \vol_{q,f}^{1/2}\;,
\end{align*}
where $n_{q,f}$ and $\vol_{q,f}$ are the unit normal vector and area, respectively, of face $f$ on surface $q$. 
This discrete definition of the SRNF is consistent with the continuous one: it preserves convergence almost everywhere of the surface $q$ and its tangent map $Tq$, and it is uniquely determined by this property.
Given two triangular meshes $q_0$ and $q_1$ with the same mesh structure, the corresponding SRNF distance is thus a sum over all faces:
\begin{equation}
\label{equ:srnf_sum}
\dist_N(q_0,q_1)^2 
= 
\sum_f \left| \srnf_{q_0,f}-\srnf_{q_1,f} \right |^2\;.
\end{equation}

\subsection{SRCF distances between triangular meshes}
Higher-order feature maps such as the SRCF do not extend uniquely by a limiting procedure from smooth surfaces to triangular meshes. 
Therefore, we resort to principles of discrete differential geometry for their definition \citep[see][]{bobenko2008discrete}. 
Recall that the SRCF of a smooth surface $q$ is defined as $\Phi_q=H_q\vol_q^{1/2}$, where $H_q$ is the vector mean curvature.
Following \citet{sullivan2008curvatures}, we define the discrete SRCV on each vertex $v$ as
\begin{equation*}
\Phi_{q,v}
\coloneqq
\frac{\frac12\sum_{e\ni v} e\times (n_1-n_2)}{\sqrt{\frac13\sum_{f\ni v}\vol_{q,f}}}\;,
\end{equation*}
where $e$ and $f$ run through all edges and faces adjacent to the vertex $v$, 
and $n_1,n_2$ are the unit-normal vectors of the two faces adjacent to the edge $e$.
Equivalently,
\begin{equation}
\Phi_{q,v}=
\frac{\frac12\sum_{f\ni v} e\times n}{\sqrt{\frac13\sum_{f\ni v}\vol_{q,f}}}\;,
\end{equation}
where $e$ is the edge opposite of $v$ in $f$, and $n$ is is the unit-normal vector of $f$.
Both formulas can be expressed equivalently by the well-known co-tangent formula \citep{sullivan2008curvatures}.
The discrete SRCF energy is defined by summing over all vertices:
\begin{equation*}
\dist_\Phi(q_0,q_1)^2
\coloneqq
\sum_v (\Phi_{q_0,v}-\Phi_{q_1,v})^2\;.
\end{equation*}

\subsection{Varifold distances between triangular meshes}
Varifold distances have a natural discretization for triangular meshes. 
We briefly recap the main elements using the notation of Section~\ref{sec:varifold}, referring to 
\citet{kaltenmark2017general} or \citet{Charon2020fidelity} for further details.
For any triangular surfaces $q_0$ and $q_1$, the varifold inner product \eqref{equ:norm_var} can be expressed as a sum over pairs of faces $f_0$ and $f_1$:
\begin{multline*}
 \langle\mu_{q_0},\mu_{q_1}\rangle_{V^*} = \sum_{f_0,f_1} \iint_{f_0 \times f_1} \rho\big(|x_0-x_1|\big) \\
 \gamma\big(n_{q_0}(x_0) \cdot n_{q_1}(x_1)\big) \vol_{q_0}(x_0) \vol_{q_1}(x_1)
\end{multline*}
Importantly, meshes $q_0$ and $q_1$ may have different topologies.
Note that the normal vector fields are constant on the domain of the double integral, i.e., on each pair of faces.
Moreover, assuming that the faces are small compared to the variation scale of the kernel $\rho$, the function $\rho(|x_0-x_1|)$ is well-approximated on the domain of the double integral by the constant $\rho(|c_{q_0,f_0}-c_{q_1,f_1}|)$, where $c_{q,f}$ is the barycenter of face $f$ within surface $q$:
\begin{equation*}
\textstyle
c_{q,f} \coloneqq \big(\sum_{v \in f} v\big)/\big(\sum_{v \in f} 1\big)
\end{equation*}
The approximation we use then writes:
\begin{multline*}
\langle\mu_{q_0},\mu_{q_1}\rangle_{V^*}
\approx 
\sum_{f_0,f_1} \rho\big(|c_{q_0,f_0}-c_{q_1,f_1}|\big)
\\
\gamma\big(n_{q_0,f_0} \cdot n_{q_1,f_1}\big) \vol_{q_0,f_0} \vol_{q_1,f_1}
\end{multline*}
This discretization easily extends to functional varifolds, as described in Section~\ref{sec:varifold}:
the varifold inner product of textured triangular surfaces $(q_0,\zeta_{q_0})$ and $(q_1,\zeta_{q_1})$ is computed as
\begin{multline*}
\langle\mu_{q_0,\zeta_{q_0}},\mu_{q_1,\zeta_{q_1}}\rangle_{V^*}
\approx 
\sum_{f_0,f_1} \rho\big(|c_{q_0,f_0}-c_{q_1,f_1}|\big)
\gamma\big(n_{q_0,f_0} \cdot n_{q_1,f_1}\big) 
\\ 
\tau\big(|\zeta_{q_0,f_0}-\zeta_{q_1,f_1}|\big)\vol_{q_0,f_0} \vol_{q_1,f_1}\;,
\end{multline*}
where $\zeta_{q,f}$ denotes the texture value attached to face $f$ of surface $q$.
If the texture is given on vertices rather than faces, one first distributes it to faces by setting
\begin{equation*}
\textstyle
\zeta_{q,f} 
\coloneqq 
\big(\sum_{v \in f} \zeta_{q,v}\big)/\big(\sum_{v \in f} 1\big)\;.
\end{equation*}

The accuracy of the above approximation can be precisely controlled in terms of the maximum diameter of faces and the modulus of continuity of the kernels $\rho$ and $\gamma$, as shown by  \citet{kaltenmark2017general}. 
The approximation can be also interpreted as replacing the continuous varifold associated to the surface with a finite sum of Dirac masses.  
One can see that the numerical complexity of the varifold distance evaluation is quadratic in the number of faces. 

\subsection{Energy computation and optimization}

For the minimization of the matching energies \eqref{equ:elastic_var} and \eqref{equ:var_elastic_var} over the vertex coordinates we use a quasi-Newton method, namely {\tt SciPy}'s implementation of the L-BFGS algorithm introduced by \citet{liu1989limited}. 
The most costly operation at each iteration of the optimization routine is the computation of the varifold distance and its gradient. 
Indeed, the computational complexity of varifold distances is quadratic in the number of faces, whereas the complexity of SRNF distances is only linear.
Thus, efficient varifold computations are critical for the overall speed of our matching algorithm. 
Thanks to their highly parallelizable structure, they are well-suited for graphics processor units (GPUs).
Our implementation leverages the Python library {\tt PyKeops}, which has recently been developed by  \citet{charlier2020kernel}. 
This library is tailored specifically for the fast evaluation of kernel reductions using {\tt CUDA}. Moreover, this library allows for automatic differentiation of all expressions. 
The SRNF distance is directly implemented in {\tt PyTorch} and therefore also allows for automatic differentiation.

\subsection{Multi-resolution}
\label{sec:multires}
\begin{figure*}
\centering
\includegraphics[width=\textwidth]{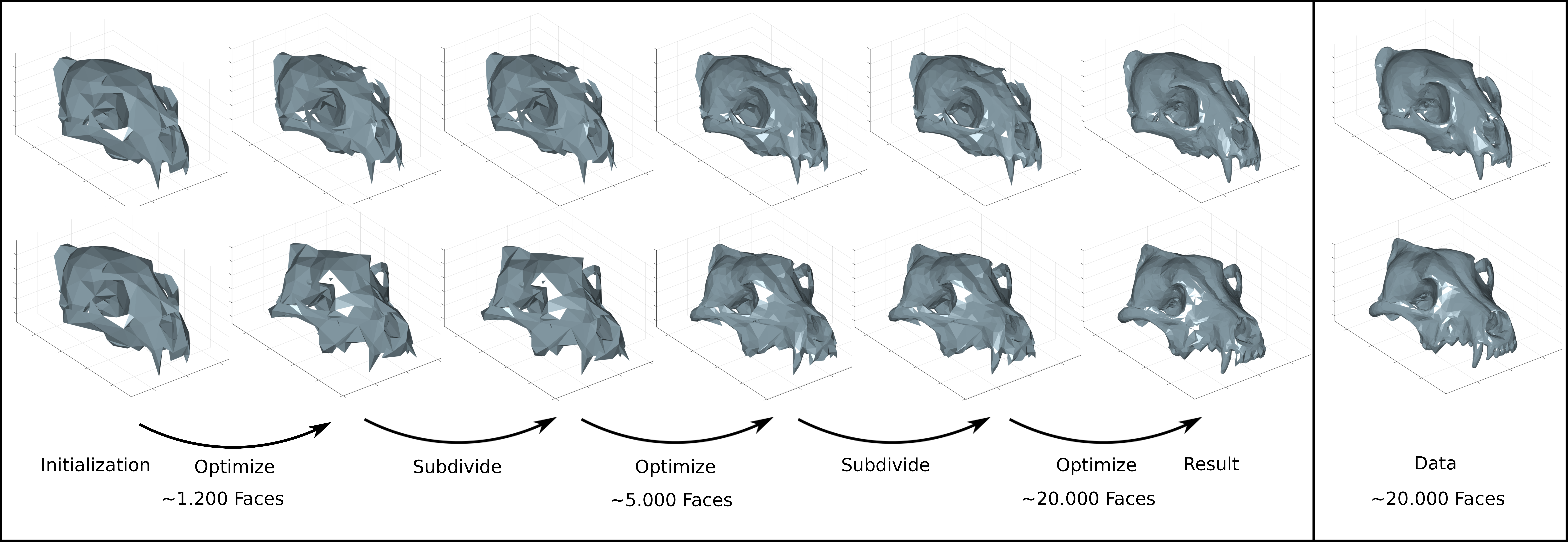}
\caption{Multi-resolution surface matching using Algorithm~\ref{alg:multires}. The data $q_0$ and $q_1$ are high-resolution triangular meshes (last column). A down-sampled version of $q_0$ is used to initialize the surfaces $\tilde q_0$ and $\tilde q_1$ (first column). These surfaces are iteratively deformed to minimize the energy functional \eqref{equ:var_elastic_var} and subdivided to achieve higher-resolution (middle columns). The result is a high-resolution point correspondence which solves the geodesic boundary value problem on shape space. 
The triangulated surfaces of the skulls are scans from the American Museum of Natural History in New York and have been obtained from 
Morphosource, a biological specimen database that has approximately 27,000 published 3D models of biological specimens.
}
\label{fig:multires}
\end{figure*}

The boundary value problem naturally lends itself to a multi-resolution implementation, where point correspondences are initially computed on a coarse grid and then iteratively refined. 
This makes it easier to find an initial guess where the optimizer for the matching energy \eqref{equ:elastic_var} or \eqref{equ:var_elastic_var} does not get trapped in a local minimum. 
Similar multi-resolution approaches have been developed by \cite{kilian2007geometric} for given point correspondences and by \cite{marin2019high} in the context of functional maps.
For simplicity, we only describe the implementation of the symmetric matching problem \eqref{equ:var_elastic_var}.
This is detailed in Algorithm~\ref{alg:multires} and illustrated in Figure~\ref{fig:multires}. 
The input to the algorithm are two triangular surfaces $q_0$ and $q_1$, which may have different mesh connectivities and topologies. 
At initialization, the variables $\tilde q_0$ and $\tilde q_1$ are set to some low-resolution surfaces with the same mesh structure. 
For instance, one may use a coarse triangulation of a sphere or some other problem-specific shape prior. 
Alternatively, one may also use a down-sampled version of $q_0$ or $q_1$, albeit at the cost of destroying the symmetry of the matching algorithm.
In any case, the initialization has to be done such that $\tilde q_0$ and $\tilde q_1$ share the same mesh structure, as this is required in the computation of SRNF distances.
Then, iteratively, the surfaces $\tilde q_0$ and $\tilde q_1$ are deformed in order to minimize the matching functional \eqref{equ:var_elastic_var} and up-sampled to higher resolutions. 
The energy minimization is implemented as described in the previous section. 
Up-sampling is implemented using a subdivision scheme. 
The easiest choice is a non-adaptive scheme, which divides each triangle into four sub-triangles, but adaptive schemes would be possible, as well. 
Functional data, if present, have to be mapped to the new mesh based on the initial fully sampled surface by e.g.\ closest point(s) interpolation.

\begin{algorithm}
\Fn{\Match{$\tilde q_0$, $\tilde q_1$, $q_0$, $q_1$}}{
  Using an iterative optimization algorithm initialized with $(\tilde q_0,\tilde q_1)$, minimize the matching energy \eqref{equ:var_elastic_var} with boundary data $(q_0,q_1)$ over all surface meshes with the same combinatorics as $(\tilde q_0,\tilde q_1)$ and return the minimizer.
} 
\Fn{\MultiResolutionMatch{$q_0$, $q_1$}}{
  Initialize $(\tilde q_0,\tilde q_1)$ at low resolution\;
  $(\tilde q_0,\tilde q_1) \longleftarrow \Match(\tilde q_0,\tilde q_1,q_0,q_1)$\;
  \For{{\rm several times}}{
     $(\tilde q_0,\tilde q_1) \longleftarrow \UpSample(\tilde q_0,\tilde q_1)$\;
     $(\tilde q_0,\tilde q_1) \longleftarrow \Match(\tilde q_0,\tilde q_1,q_0,q_1)$\;
  }
  \KwRet{$(\tilde q_0,\tilde q_1)$}
} 
\caption{Multi-resolution matching}
\label{alg:multires}
\end{algorithm}

\subsection{Inversion of the SRNF map}
\label{sec:srnf_inversion}

\begin{figure*}
\centering
\def\svgwidth{\linewidth}
\begingroup%
  \makeatletter%
  \providecommand\color[2][]{%
    \errmessage{(Inkscape) Color is used for the text in Inkscape, but the package 'color.sty' is not loaded}%
    \renewcommand\color[2][]{}%
  }%
  \providecommand\transparent[1]{%
    \errmessage{(Inkscape) Transparency is used (non-zero) for the text in Inkscape, but the package 'transparent.sty' is not loaded}%
    \renewcommand\transparent[1]{}%
  }%
  \providecommand\rotatebox[2]{#2}%
  \ifx\svgwidth\undefined%
    \setlength{\unitlength}{1554.04985082bp}%
    \ifx\svgscale\undefined%
      \relax%
    \else%
      \setlength{\unitlength}{\unitlength * \real{\svgscale}}%
    \fi%
  \else%
    \setlength{\unitlength}{\svgwidth}%
  \fi%
  \global\let\svgwidth\undefined%
  \global\let\svgscale\undefined%
  \makeatother%
  \begin{picture}(1,0.25989007)%
    \put(0,0){\includegraphics[width=\unitlength,page=1]{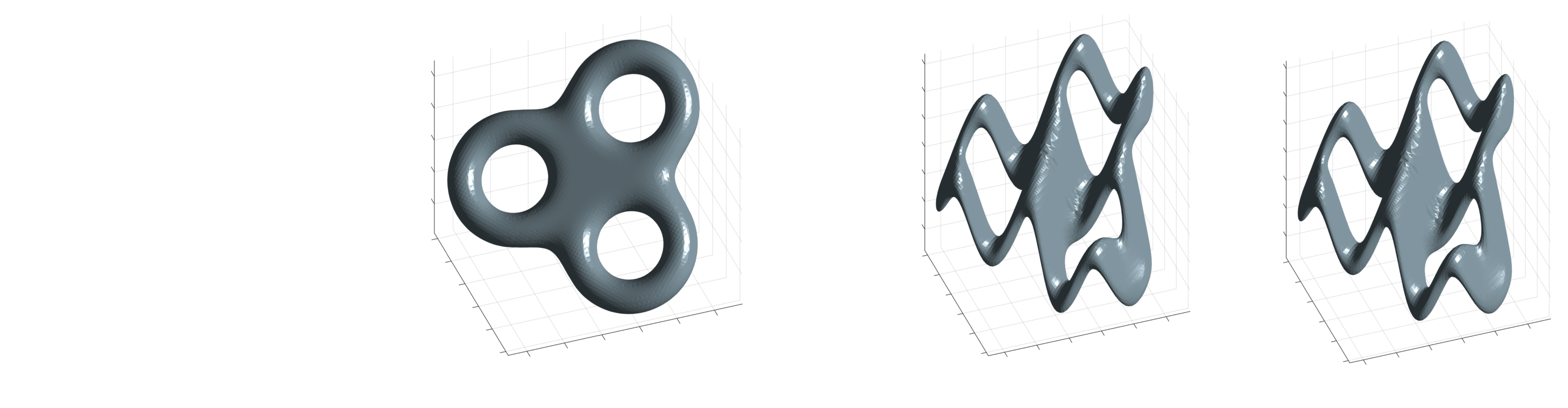}}%
    \put(0.49443211,0.11662487){\color[rgb]{0,0,0}\makebox(0,0)[lb]{\smash{{\tiny Optimize}}}}%
    \put(0.8672307,0.00976034){\color[rgb]{0,0,0}\makebox(0,0)[lb]{\smash{Ground Truth}}}%
    \put(0.3350493,0.01140962){\color[rgb]{0,0,0}\makebox(0,0)[lb]{\smash{Initial Guess}}}%
    \put(0.66258768,0.01074809){\color[rgb]{0,0,0}\makebox(0,0)[lb]{\smash{Result}}}%
    \put(0,0){\includegraphics[width=\unitlength,page=2]{inversion3.pdf}}%
    \put(0.08194356,0.01103392){\color[rgb]{0,0,0}\makebox(0,0)[lb]{\smash{SRNF}}}%
    \put(0,0){\includegraphics[width=\unitlength,page=3]{inversion3.pdf}}%
  \end{picture}%
\endgroup%
\caption{Approximate inversion of the SRNF-map. Given the SRNF (left) of an unknown surface (right), optimization of the energy functional \eqref{equ:energy_inversion} determines a surface (middle) whose SRNF closely matches the given one. The algorithm works reliably for high-genus and high-resolution surfaces (here with approximatively $25,000$ faces).}
\label{fig:SRNF-inversion1}  
\end{figure*}

\begin{figure*}
\centering
\captionsetup[subfigure]{labelformat=empty}
\subfloat[$t=0$\\ $\tilde q_0$]{\includegraphics[width=0.15\textwidth]{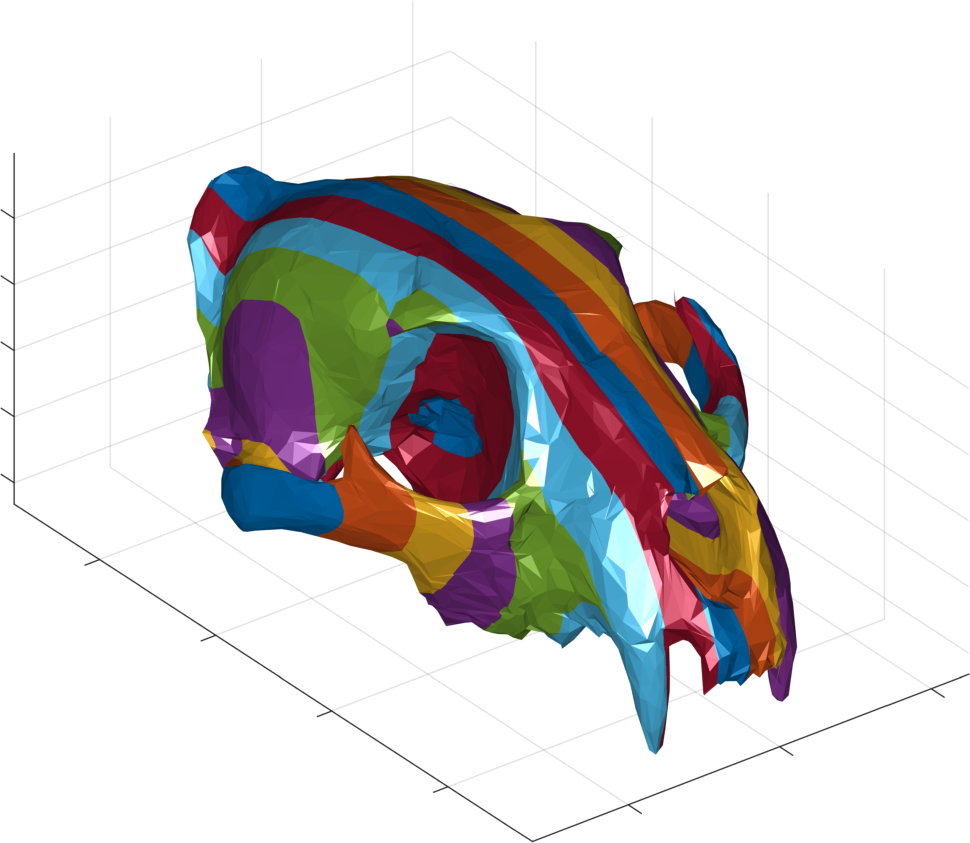}}
\hspace{.15cm}
\subfloat[$t=0.2$]{\includegraphics[width=0.15\textwidth]{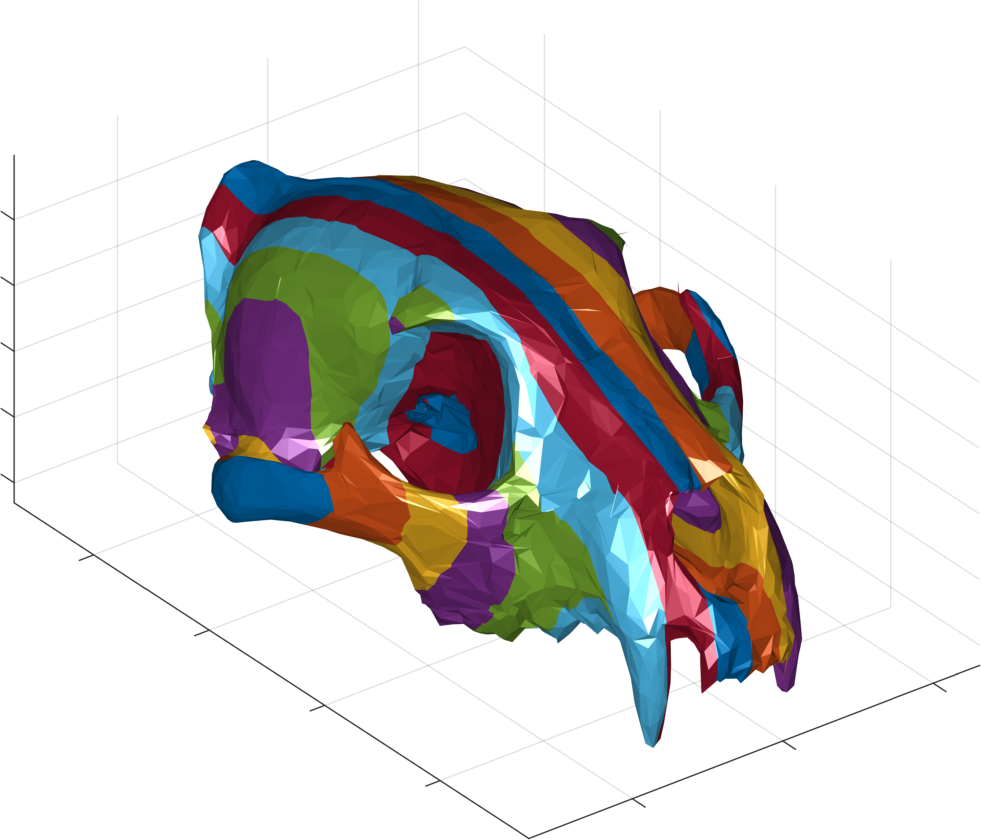}}
\hspace{.15cm}
\subfloat[$t=0.5$]{\includegraphics[width=0.15\textwidth]{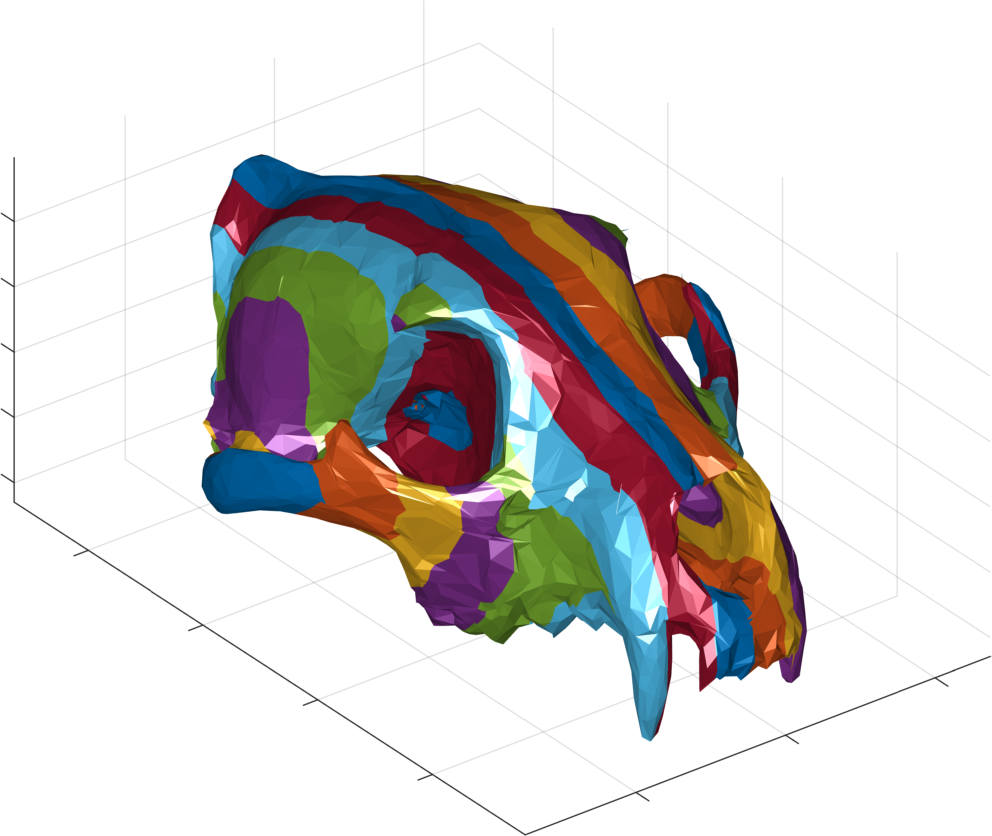}}
\hspace{.15cm}
\subfloat[$t=0.6$]{\includegraphics[width=0.15\textwidth]{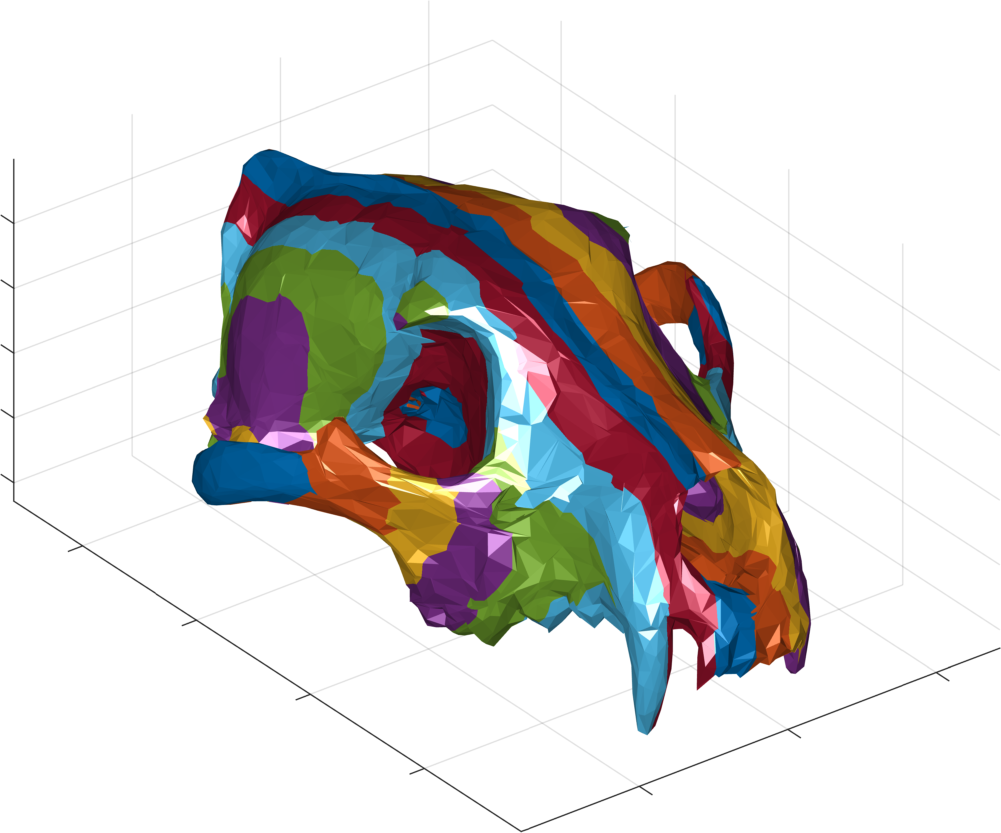}}
\hspace{.15cm}
\subfloat[$t=0.8$]{\includegraphics[width=0.15\textwidth]{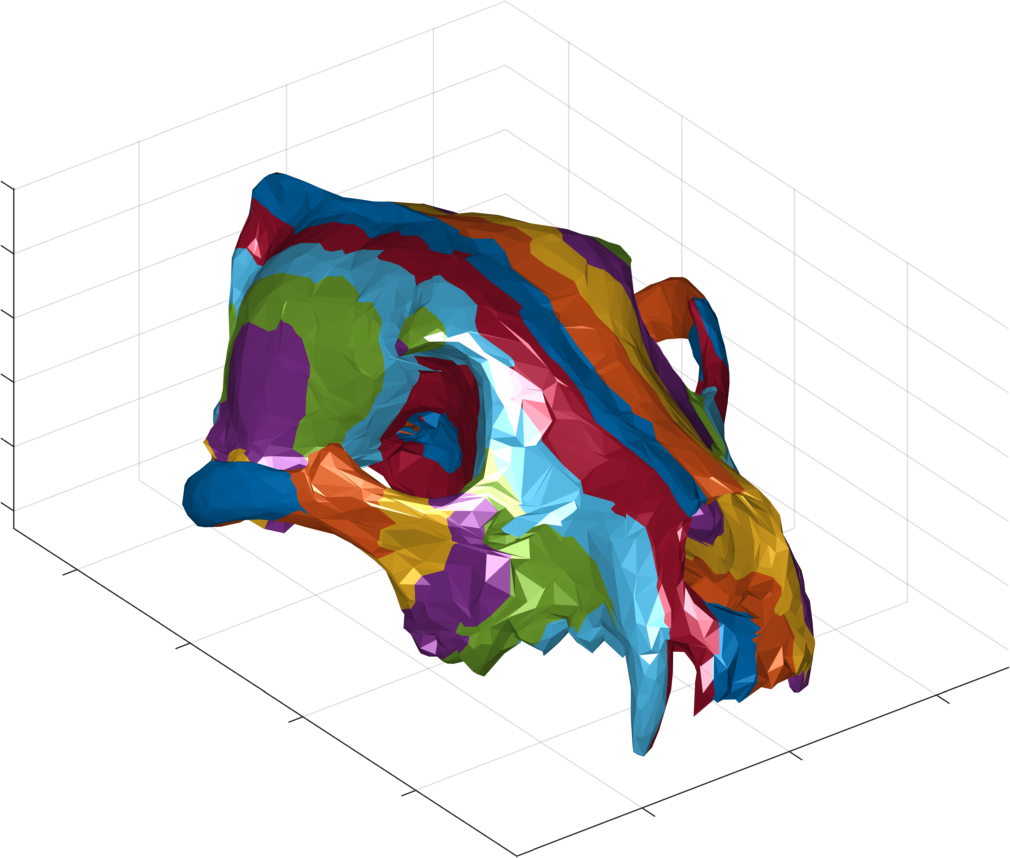}}
\hspace{.15cm}
\subfloat[$t=1$\\ $\tilde q_1$]{\includegraphics[width=0.15\textwidth]{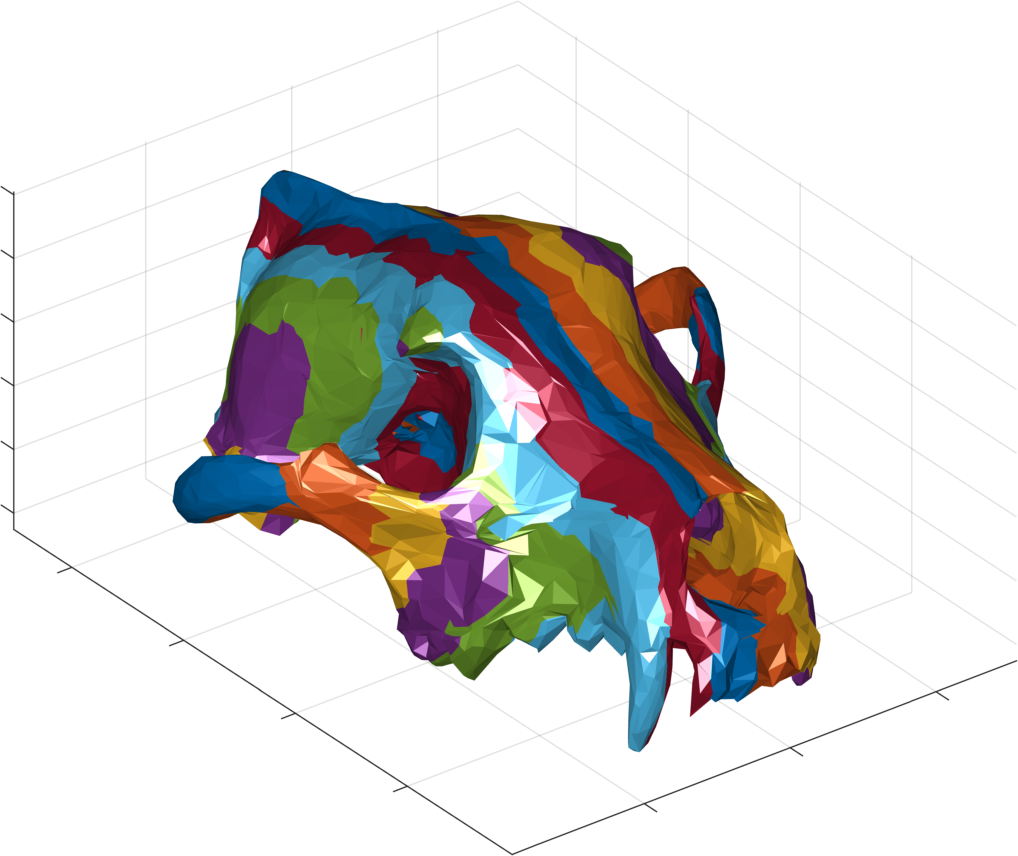}}
\caption{Computation of geodesics using Algorithm~\ref{alg:geodesics}. The geodesic is obtained by matching the given surfaces $q_0$ and $q_1$ using Algorithm~\ref{alg:multires}, linearly interpolating the SRNFs of the matched surfaces $\tilde q_0$ and $\tilde q_1$, and inverting the SRNF map. The color-coding shows that all surfaces along the geodesic are parametrized consistently.}
\label{fig:geodesic_skull1}  
\end{figure*}

Recall that Algorithm~\ref{alg:multires} computes the initial and final values of a geodesic between the given shapes.
We next describe how to obtain intermediate values of this geodesic.
These are of interest in many applications and describe the optimal elastic deformation between the shapes. 
Our algorithm is an extension and modification of \citet{laga2017numerical}.
Pseudo-code for the algorithm is shown in Algorithm~\ref{alg:geodesics}.
The starting point is the observation that geodesics in the feature space of the SRNF map are straight lines. 
Thus, given the output $\tilde q_0$ and $\tilde q_1$ of Algorithm~\ref{alg:multires}, this geodesic in feature space can be computed by linearly interpolating the SRNFs of $\tilde q_0$ and $\tilde q_1$. 
To find surfaces corresponding to these interpolated SRNFs, one has to invert the SRNF map. 
Of course, inversion in the strict sense is impossible because the SRNF map is neither injective nor surjective \citep{klassen2019closed}.
In particular, the linear interpolation of two SRNFs may not be the SRNF of any surface, and if such a surface exists it may be non-unique. 
We circumvent this issue by recasting it as an optimization problem:
given a function $\tilde N\colon M\to \mathbb R^3$, which is constant on each face $f$ of the triangular surface $M$, find a triangular surface mesh $q$ which minimizes the energy functional
\begin{equation}
\label{equ:energy_inversion}
\int_M | \srnf_q- \tilde N|^2 =
\sum_f | \srnf_{q,f}- \tilde N_f|^2\;.
\end{equation}
Similarly as in the matching problem, we minimize this energy functional using an L-BFGS method, where the gradient is computed by automatic differentiation in {\tt PyTorch}. 
This improves upon the optimization routine of \cite{laga2017numerical}, where no information about the Hessian is used.
Despite the analytical difficulties pointed out in \cite{klassen2019closed}, this optimization problem turns out to be rather well-behaved in our experiments. 
For instance, the optimizer is able to recover a surface from its SRNF even when the initial guess is quite far off, as shown in  Figure~\ref{fig:SRNF-inversion1}. 
Moreover, one actually disposes of quite accurate initial guesses in the context of matching problems, where a linear path in SRNF-space has to be inverted. 
An example of a resulting optimal elastic deformation is shown in Figure~\ref{fig:geodesic_skull1}.
A similar strategy can be used for the approximate inversion of the SRCF map, where one augments the energy functional~\eqref{equ:energy_inversion}
by the $L^2$-difference between the SRCFs. 
\begin{algorithm}
\Fn{\InverseSRNF{$\tilde N$}}{
Minimize the energy \eqref{equ:energy_inversion} over all triangular meshes $q$ with the same combinatorics as $\tilde N$ and return the minimizer.
}
\Fn{\GeodesicInterpolation{$t$, $q_0$, $q_1$}}{
  $(\tilde q_0,\tilde q_1) \longleftarrow \MultiResolutionMatch(q_0,q_1)$ \;
  $\tilde N \longleftarrow (1-t)\srnf_{\tilde q_0} + t \srnf_{\tilde q_1}$ \;
  $q \longleftarrow $ \InverseSRNF($\smash{\tilde N}$)\;
  \KwRet{$q$}
  }
\caption{Computation of geodesics}
\label{alg:geodesics}
\end{algorithm}

\section{Numerical results}
\label{sec:experiments}
In this final section we demonstrate various features of our algorithm through numerical simulations. 

\subsection{Choice of parameters}
There are several parameters to be set. 
Specifically, these are the kernels in the varifold fidelity metric and the coefficient $\lambda$ which weighs the contribution of the SRNF or SRCF term relative to the varifold terms in the matching energy. 
In all our simulations, we used the Gaussian kernel 
\begin{equation*}
\rho(|x_1-x_2|)
\coloneqq
\exp(-|x_2-x_1|^2/\sigma^2)
\end{equation*} of scale $\sigma$ on $\mathbb R^3$ and the Binet kernel 
\begin{equation*}
\gamma(n_1 \cdot n_2) 
\coloneqq 
(n_1 \cdot n_2)^2
\end{equation*} on $S^2$. 
Our implementation allows for many other choices of kernels, and we refer to \cite{kaltenmark2017general} for a thorough discussion of the effects that this can have in shape matching problems. 
The kernel scale $\sigma$ and weight $\lambda$ have to be selected in accordance with the data. 
This can be done in an adaptive fashion in combination with the multi-re\-so\-lu\-tion approach described earlier. 
Namely, we typically initialize the kernel scale $\sigma$ to a value around a tenth of the shape's diameter and decrease it throughout the successive runs of the multi-re\-so\-lu\-tion algorithm. On the opposite, the relaxation parameter $\lambda$ is increased after each run in order to enforce a closer and closer matching to the target shape. We also point to the demo scripts provided in the {\tt github} package for examples of parameter setting. 

\subsection{Choice of elastic energy}
\begin{figure*}[h]
\captionsetup[subfigure]{labelformat=empty}
\centering
\subfloat[$t=0$\\$\tilde q_0$]{\includegraphics[width=0.15\textwidth]{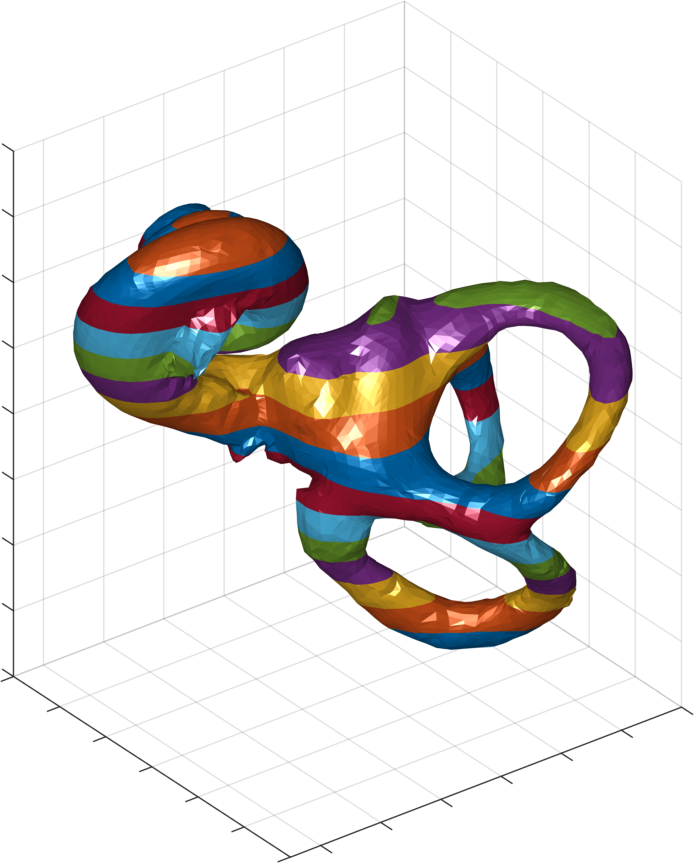}}
\hspace{.15cm}
\subfloat[$t=0.2$]{\includegraphics[width=0.15\textwidth]{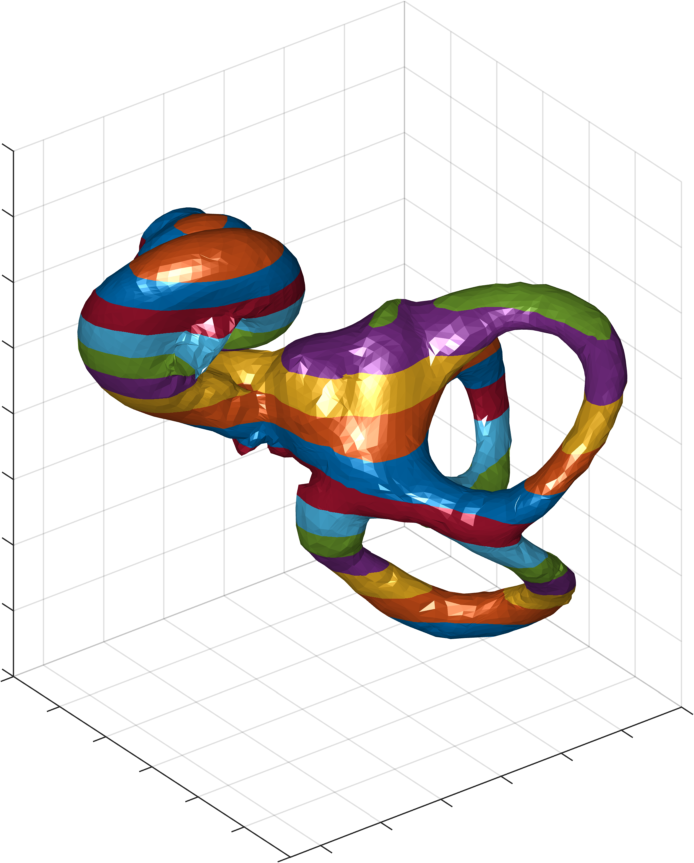}}
\hspace{.15cm}
\subfloat[$t=0.4$]{\includegraphics[width=0.15\textwidth]{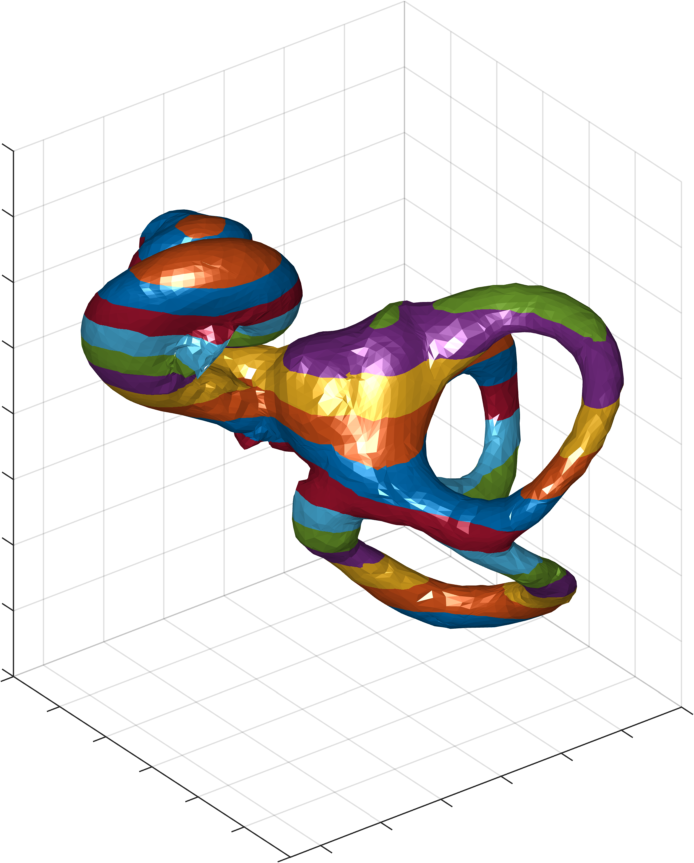}}
\hspace{.15cm}
\subfloat[$t=0.6$]{\includegraphics[width=0.15\textwidth]{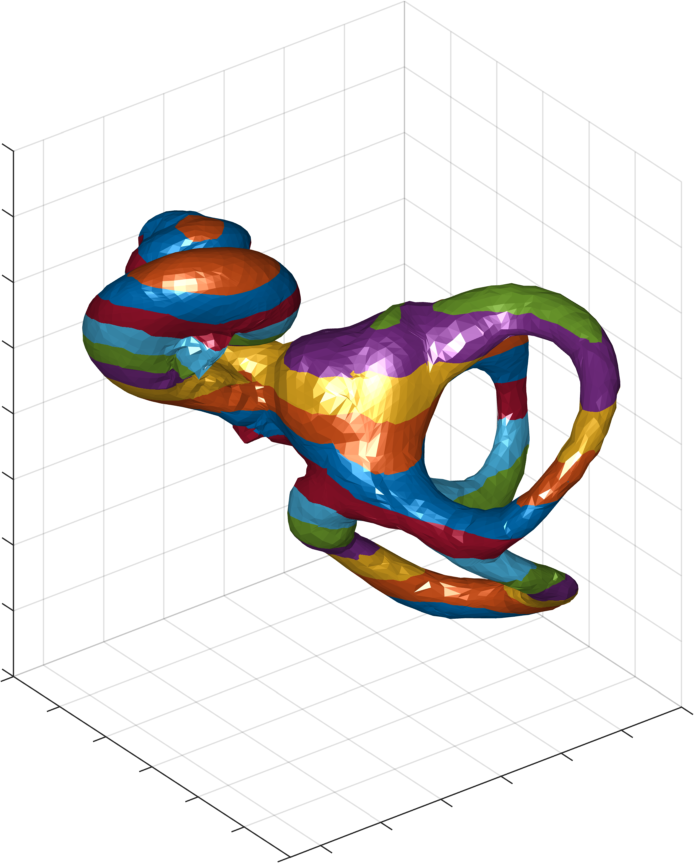}}
\hspace{.15cm}
\subfloat[$t=0.8$]{\includegraphics[width=0.15\textwidth]{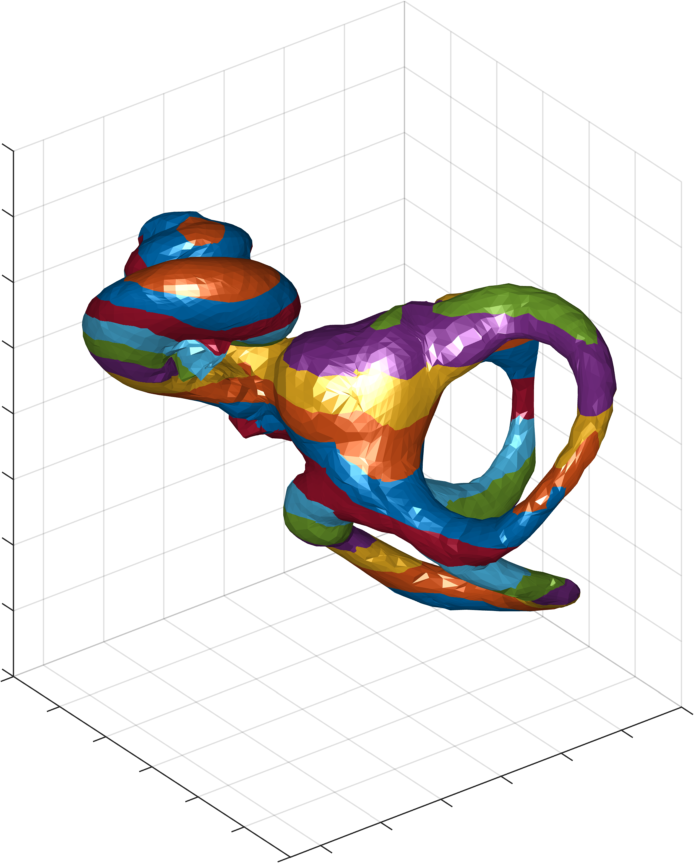}}
\hspace{.15cm}
\subfloat[$t=1$\\$\tilde q_1$]{\includegraphics[width=0.15\textwidth]{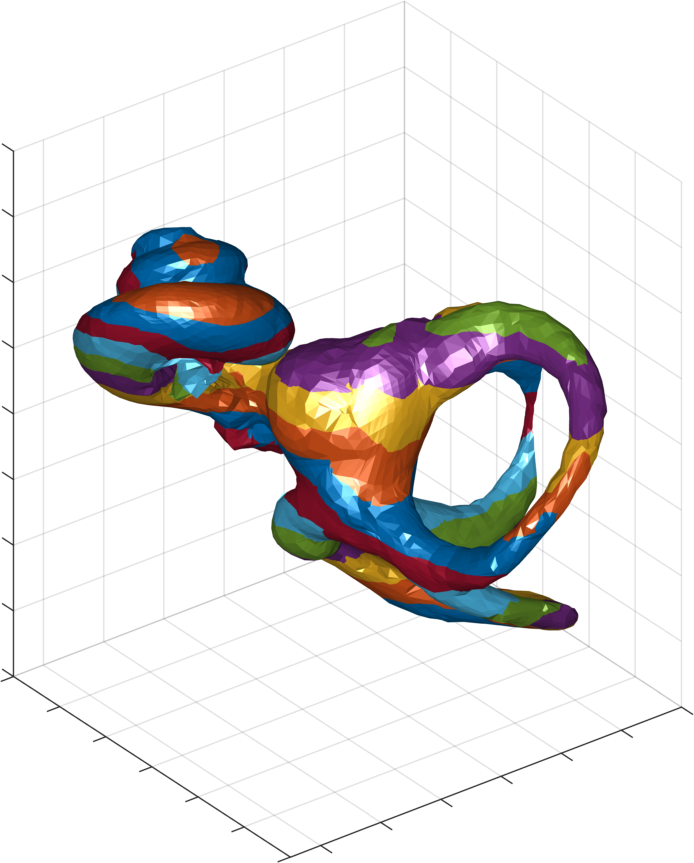}}
\caption{Matching high-genus surfaces. Two cochleas (left, right) are given as triangular meshes without any point correspondence between them. Algorithms~\ref{alg:multires} and~\ref{alg:geodesics} determine a consistent parametrization (color-coded) for the two surfaces and the geodesic in-between (middle).}
\label{fig:cochlea}       
\end{figure*}
We used either the SRNF distance or a combination of the SRNF and SRCF distances as deformation energies in our numerical experiments. 
SRNF distances reliably led to good results and stable optimization routines and were used for most of our numerical simulations. 
SRCF distances alone had the tendency to create numerical instabilities. 
Combined with SRNF distances, according to our limited numerical evidence, these instabilities disappeared.

As expected, the choice of deformation energy influences the resulting point correspondences and distances, and thus impacts the statistical analysis on the space of shapes.
This is demonstrated in a toy example in Figure~\ref{fig:cups_glasses}.
In future works it will be interesting to choose the deformation energy in a data-driven way, as suggested by \citet{needham2020simplifying} for elastic metrics on spaces of curves and by \citet{niethammer2019metric} for diffeomorphic metrics.

\subsection{Run time}
\label{sec:run_time}
As an indication of the algorithm's practical performance, we show in Table~\ref{tab:computationtime} the average run time per iteration of the L-BFGS optimization scheme for the (non-symmetric) surface matching algorithm. Typically, in those experiments, the algorithm required around 300 iterations to reach a relative error of less than $0.1\%$. All those computation times were obtained on an Intel i7-9700K processor with 3.6 GHz clock rate and a 2018 Nvidia GeForce RTX 2080 Ti with 11GB memory. We ran the matching between two surfaces with a range of different mesh sizes, both for the SRNF and the combined SRNF-SRCF matching procedures. 

To further put those results into perspective, we also performed similar tests for another well-established surface registration framework based on the Large Deformation Diffeomorphic Metric Mapping (LDDMM) model of \cite{beg2005computing}. This approach shares some similarity with the one presented in this paper in that both solve inexact matching problems with varifold distances as relaxation terms, the essential difference being that the LDDMM algorithm estimates a dense and smooth deformation of the whole ambient space to transform the source shape. This typically induces a higher numerical complexity compared to the SRNF framework that directly optimizes over the deformed source shape, as evidenced by run times per iteration approximately $10$ times larger. For the computation of the diffeomorphic flow, we used a Ralston integrator with $10$ time steps. We emphasize that, in order to make those comparisons meaningful, both the SRNF and LDDMM models were implemented in {\tt Python} based on the {\tt Scipy} implementation of the L-BFGS algorithm and rely on {\tt CUDA} subroutines as well as automatic differentiation through the use of the same {\tt PyTorch} and {\tt PyKeops} libraries.

\begin{table}[h!]
\begin{center}
\begin{tabular}{@{}lccc@{}} 
 \toprule
  & \multicolumn{2}{c}{Time per iteration (s)}\\ \cmidrule(l){2-3}
  Faces & SRNF  &  SRNF \& SRCF & LDDMM \\
  \midrule
 1,000 & 0.007  & 0.01 & 0.11\\
  10,000 &0.016 &0.018  & 0.22 \\
 50,000 &0.12& 0.12 & 1.1\\
  200,000&1.73 &1.84 & 16.1\\
  \bottomrule
\end{tabular}
\end{center}
\caption{Computation times for a single iteration of the L-BFGS algorithm, which is here applied to minimize the non-symmetric matching energy \eqref{equ:elastic_var}, for different resolutions of the mesh. For comparison, we also provide the similar time per iteration for an alternative surface registration approach based on a diffeomorphic model (LDDMM).}
\label{tab:computationtime}
\end{table}

\subsection{High-genus surfaces}
All parts of the algorithm, from varifold distances to SRNF distances and SRNF inversion, are implemented for triangular meshes. 
Therefore, there are no topological restrictions on the surfaces to be matched. 
For instance, it is possible to match high-genus surfaces such as skulls or cochleas, as demonstrated in Figures~\ref{fig:geodesic_skull1} and \ref{fig:cochlea}.
In these examples, the surfaces $q_0$ and $q_1$ to be matched are given by triangular meshes with different combinatorics, and no point correspondences are known a priori.
The matching algorithm is initialized with a down-sampled version of $q_0$, as described in Section~\ref{sec:multires}.
Accordingly, the resulting interpolated surfaces have the same genus as $q_0$.
The deformation energy, which was used in these examples, is the SRNF energy.
The multi-resolution matching algorithm converged in less than 20 seconds for both examples. 
The resulting point correspondences are color-coded in the figures.

\subsection{Topological inconsistencies}
\begin{figure}
\centering
\includegraphics[width=0.45\textwidth]{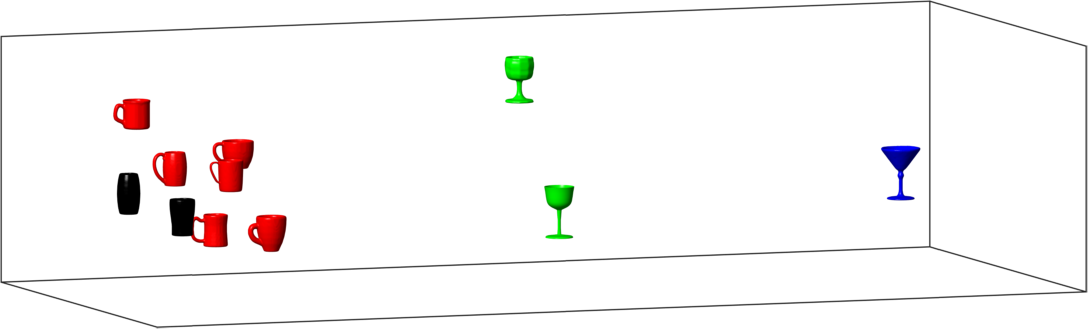}
\includegraphics[width=0.45\textwidth]{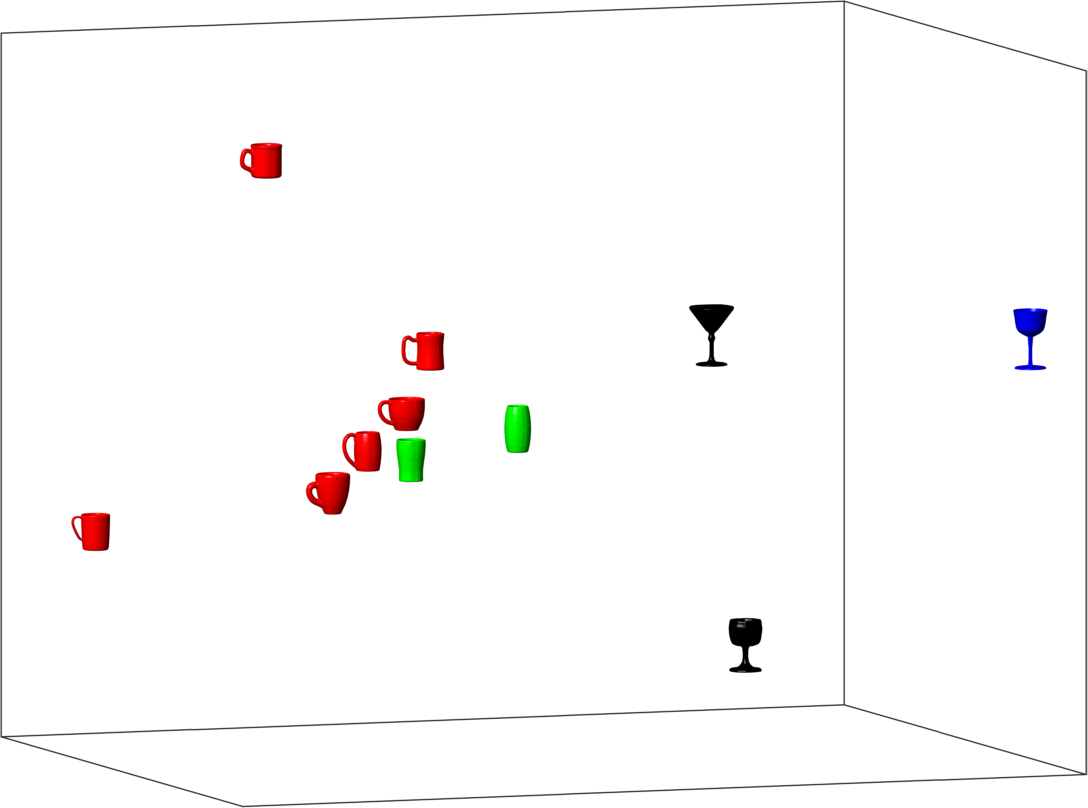}
\caption{Matching surfaces with different topologies. The cups are positioned according to multi-dimensional scaling based on the matrix of pairwise elastic-varifold discrepancies, and are colored according to an automatic distance-based clustering algorithm. Positions and colors consistently indicate that the elastic-varifold distance is able to quantify and detect both topological and geometric variations. The 
distances were computed using only SRNFs (top) or SRNFs together with SRCFs (bottom).}
\label{fig:cups_glasses}
\end{figure}

\begin{figure}
\centering
\includegraphics[width=0.15\textwidth]{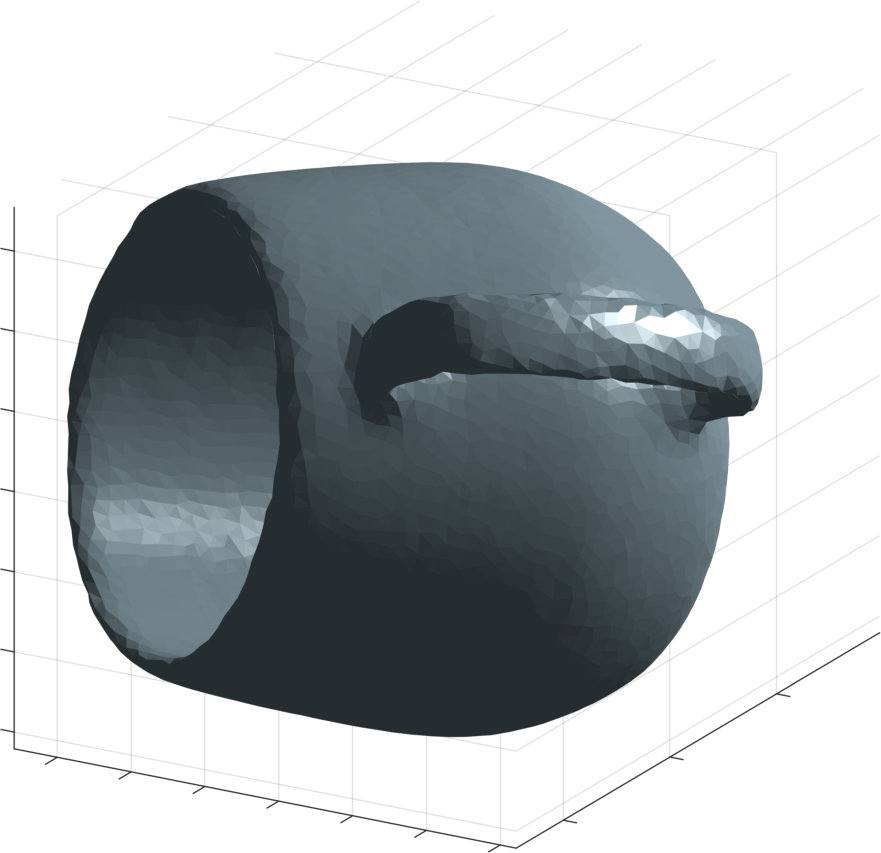}
\hspace{1cm}
\includegraphics[width=0.15\textwidth]{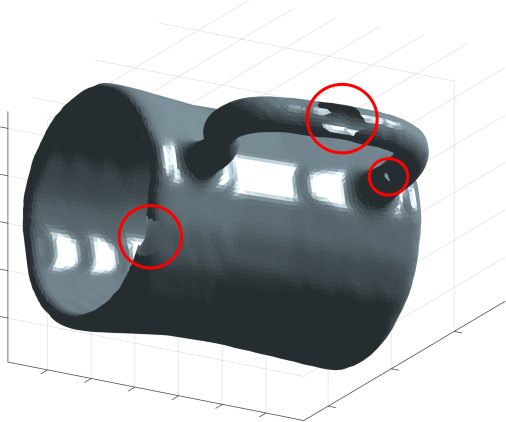}\\
\includegraphics[width=0.15\textwidth]{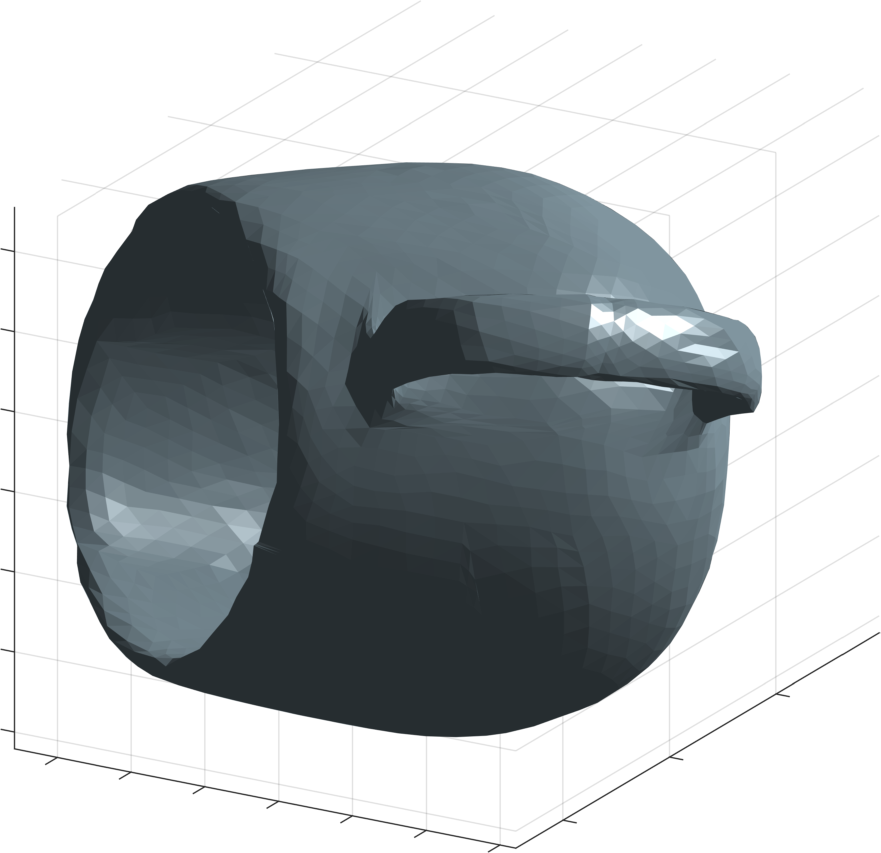}\hspace{1cm}
\includegraphics[width=0.15\textwidth]{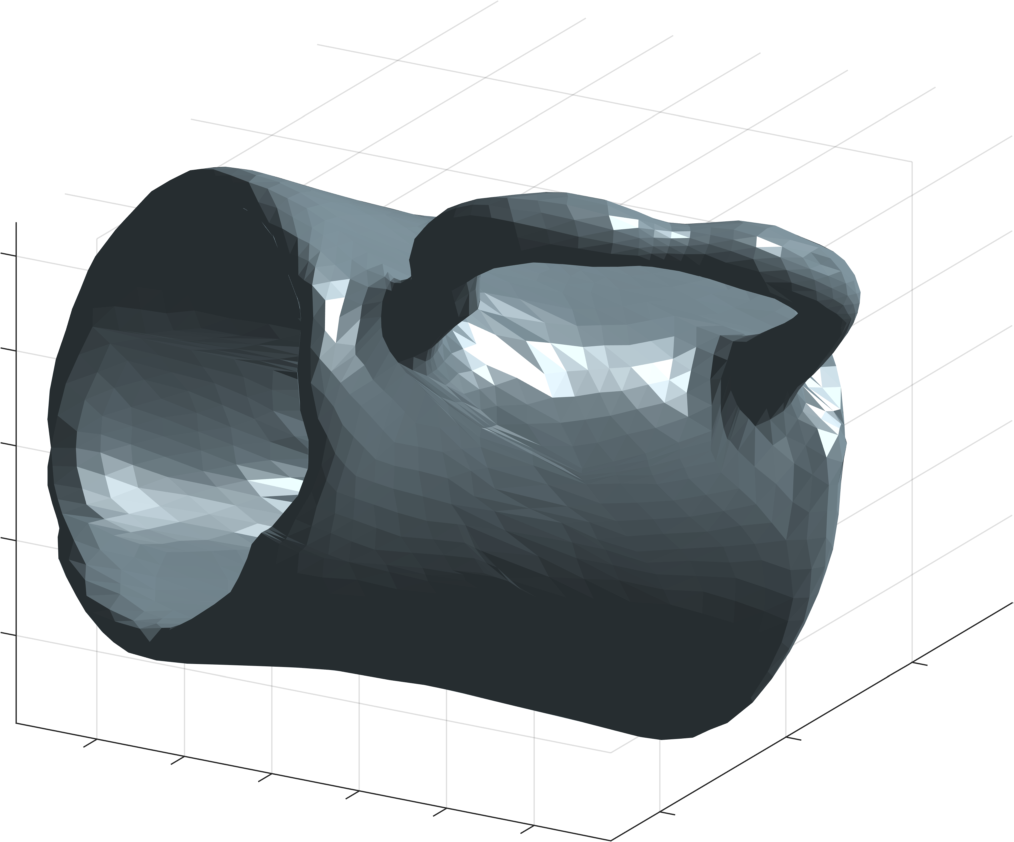}
\caption{Matching surfaces with topological noise. One of the two cups to be matched (top) has several artifacts, including a broken-off piece at the top and some holes in the triangulation at the handle. The matching algorithm is initialized by a surface without these artifacts and produces a decent fit (bottom).}
\label{fig:cups_broken}
\end{figure}

Our algorithm can handle surfaces with different topologies. 
This would not be possible in a pure elastic-mat\-ching approach, where the surfaces are required to be homotopic, see e.g. \cite{jermyn2017elastic}. This is possible in in our approach thanks to the combination with varifold distances. 
The matching energy \eqref{equ:var_elastic_var}, which is a sum of elastic and varifold terms, quantifies both geometric and topological differences and can therefore be used as a measure of discrepancy between shapes of arbitrary topology. 
For example, as shown in Figure~\ref{fig:cups_glasses}, it is possible to compare cups with handles to glasses without handles. 
In this comparison, the presence or absence of handles as well as differences in the shape of the glasses or cups are all taken into account, as revealed by clustering based on the matrix of pairwise elastic-varifold discrepancies. 

The initialization of the matching algorithm involves an important choice to be made, namely, fixing a topology and mesh structure for the surfaces $\tilde q_0$ and $\tilde q_1$ (see Algorithm~\ref{alg:multires}).
In our experiments we initialized these surfaces once with the topology of the source and once with the topology of the target  and used the average of the two resulting distances as our final discrepancy measure. 

\subsection{Topological noise}
Handling topological noise is a further application of the algorithm's capability of dealing with different topologies. 
An example is presented in Figure~\ref{fig:cups_broken}, which shows enlarged pictures of two of the cups in Figure~\ref{fig:cups_glasses}.  
One of these cups has several topological artifacts: a solid piece at the top of the cup has broken off, and several triangles are missing in the mesh describing the cup's handle. 
Nevertheless, the cup can be matched to a similar cup without these artifacts.
Of course, for topological reasons, the match cannot be exact, but it nevertheless provides a very good approximation without the missing triangles.
 
\begin{figure*}
\centering
\captionsetup[subfigure]{labelformat=empty}
\subfloat[Source $q_0$]{\fbox{\includegraphics[width=0.1\textwidth]{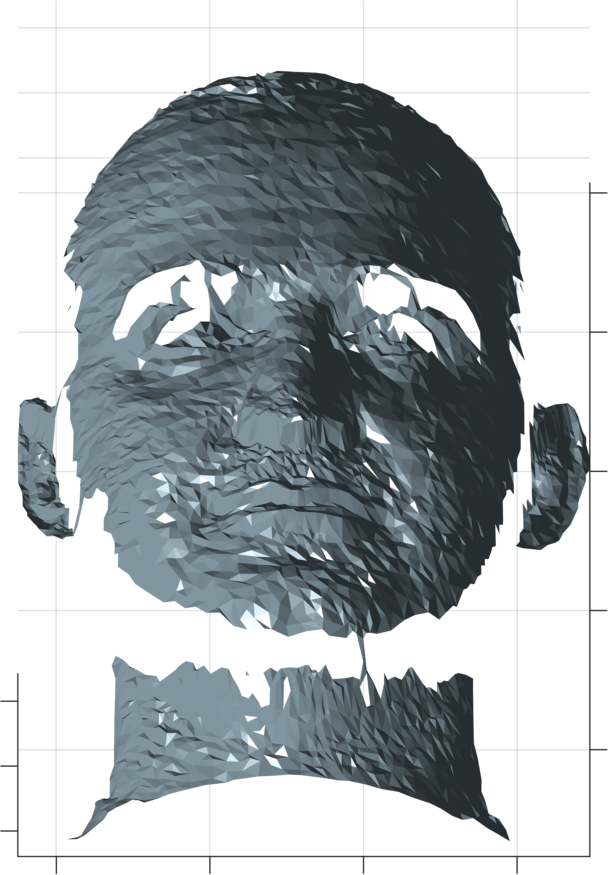}}}
\hspace{.15cm}
\subfloat[$t=0$]{\includegraphics[width=0.1\textwidth]{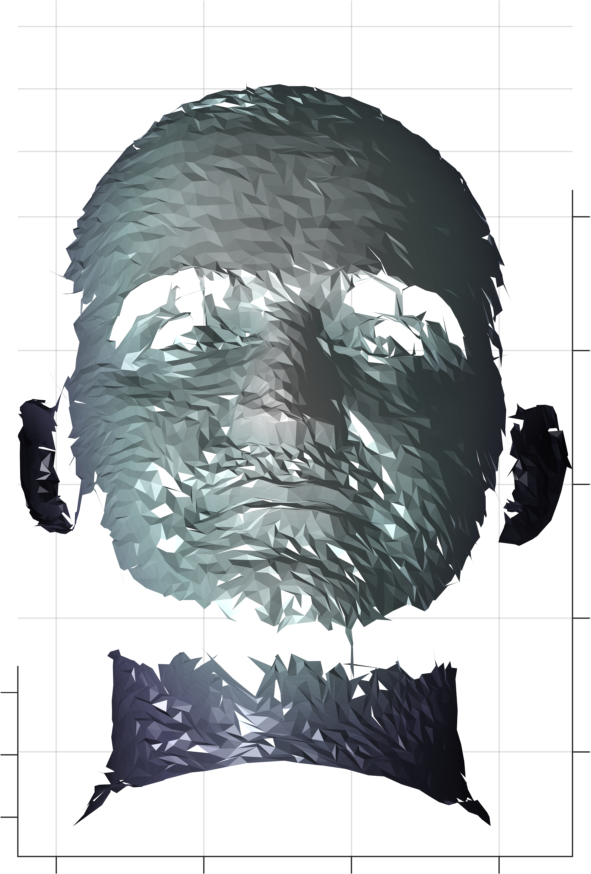}}
\hspace{.15cm}
\subfloat[$t=0.2$]{\includegraphics[width=0.1\textwidth]{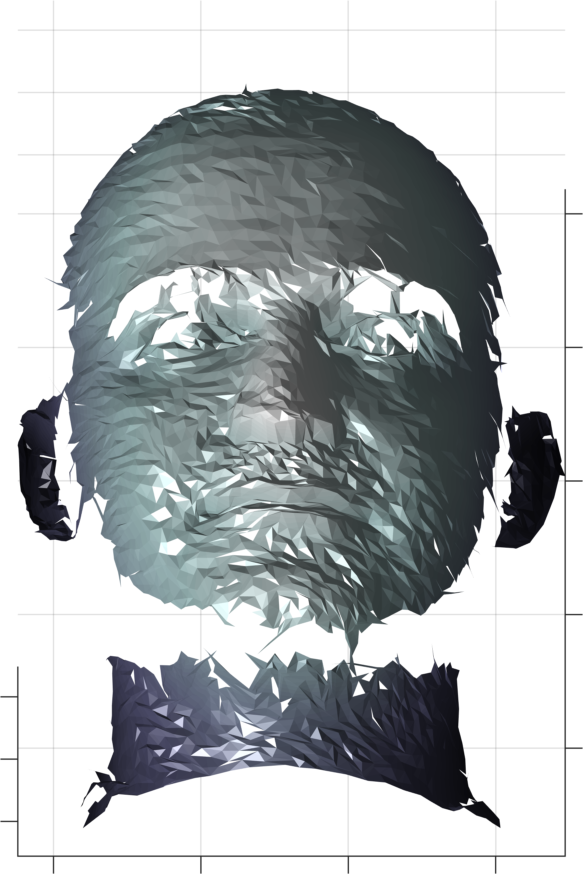}}
\hspace{.15cm}
\subfloat[$t=0.4$]{\includegraphics[width=0.1\textwidth]{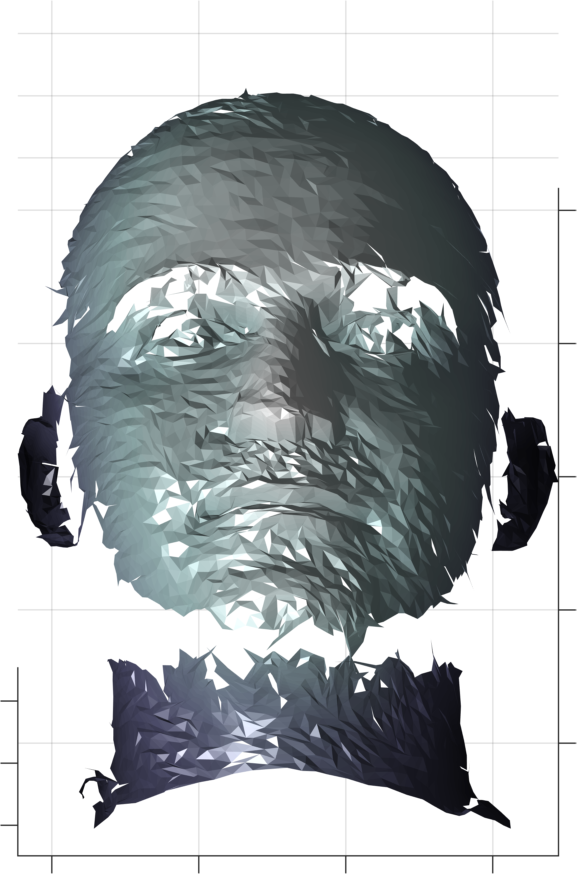}}
\hspace{.15cm}
\subfloat[$t=0.6$]{\includegraphics[width=0.1\textwidth]{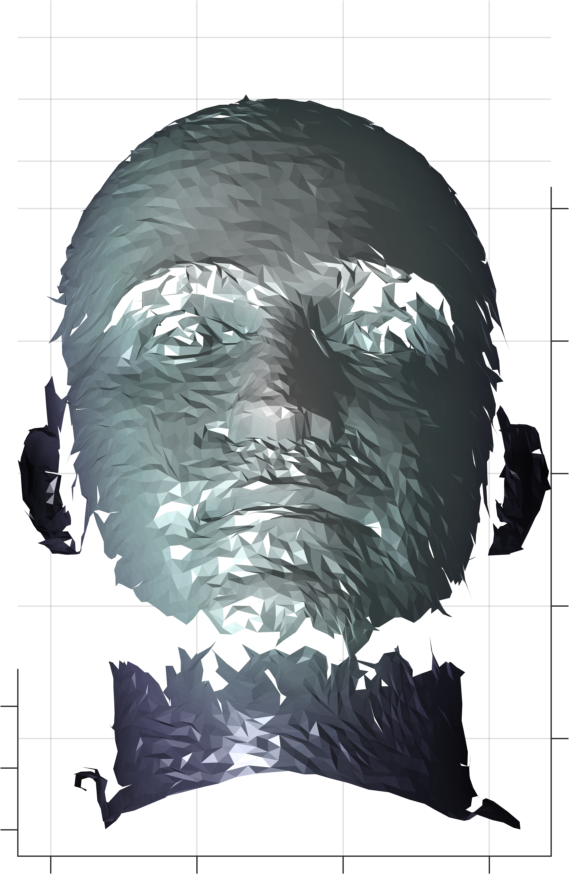}}
\hspace{.15cm}
\subfloat[$t=0.8$]{\includegraphics[width=0.1\textwidth]{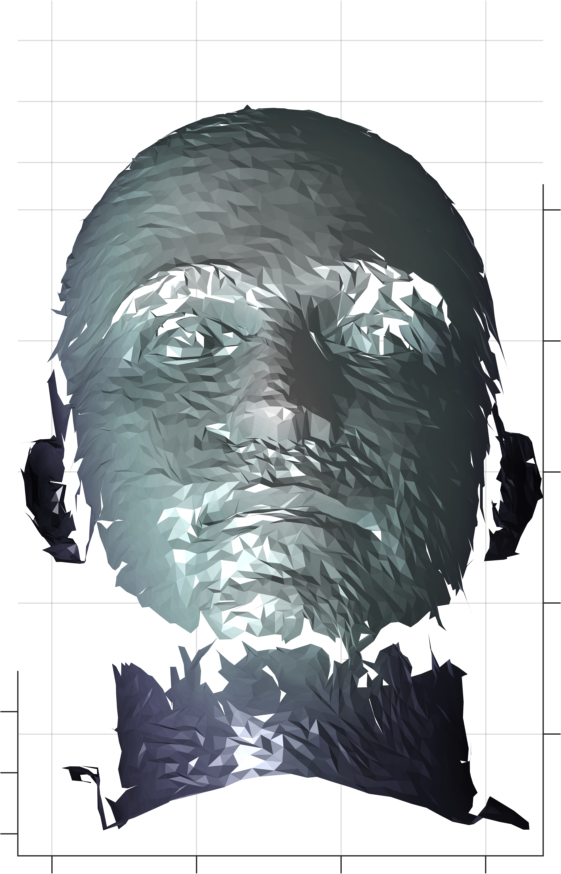}}
\hspace{.15cm}
\subfloat[$t=1$]{\includegraphics[width=0.1\textwidth]{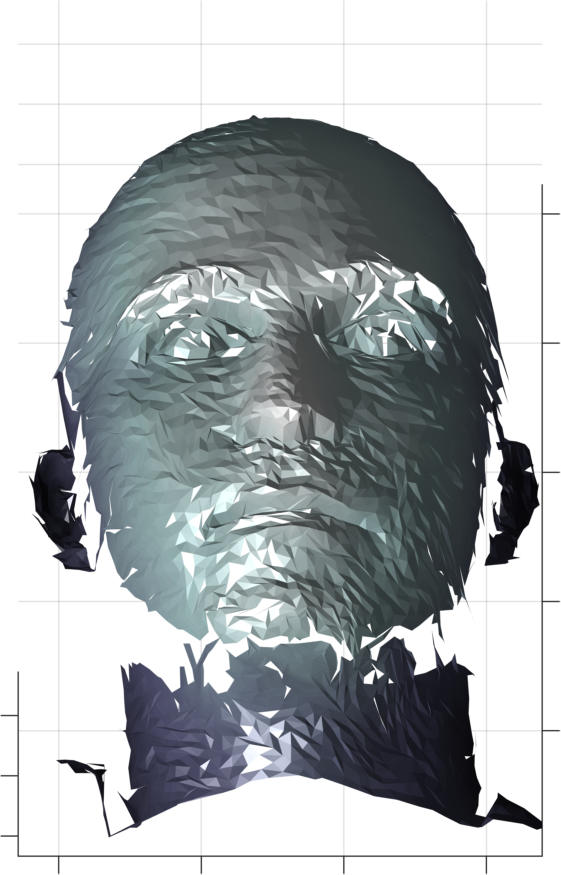}}
\hspace{.15cm}
\subfloat[Target $q_1$]{\fbox{\includegraphics[width=0.1\textwidth]{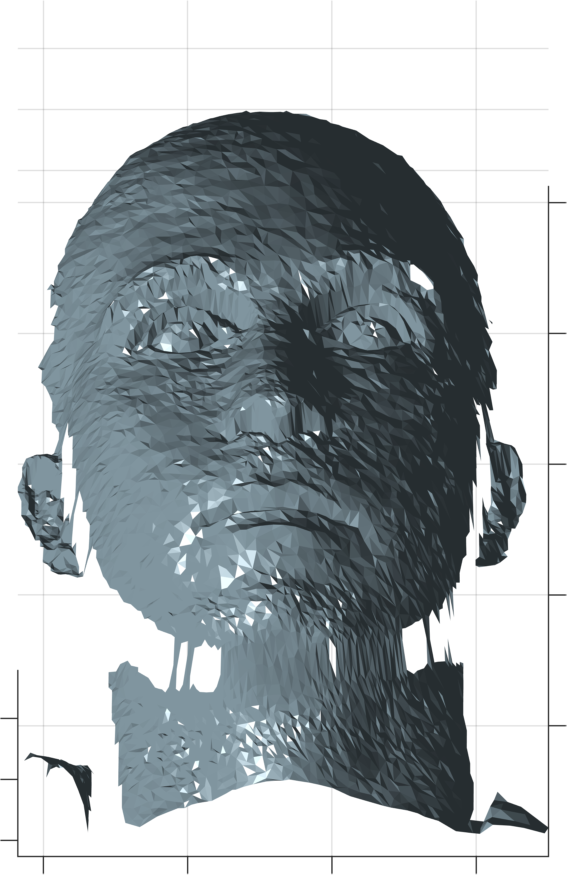}}}
\caption{Matching surfaces with topological noise. The facial scanning technology produced meshes (left, right) of varying degrees of incompleteness. Nevertheless, these meshes are matched correctly, as shown by the geodesic interpolation between them (middle).}
\label{fig:faces}
\end{figure*}

Topological artifacts are often created by 3D scanning technology, as e.g.\ in the case of the facial scans in Figure~\ref{fig:faces}. 
In this case, both facial surfaces to be matched suffer from topological noise. 
The first surface has more missing areas than the second one and is used to initialize the matching algorithm. 
The resulting geodesic interpolation gradually deforms the first face into the second one while filling out the missing areas.  

\subsection{Functional shapes}
\label{sec:functional}

\begin{figure*}
\centering
\captionsetup[subfigure]{labelformat=empty}
\begin{minipage}{0.13\textwidth}
\subfloat[Source $q_0$]{\fbox{\includegraphics[width=\textwidth]{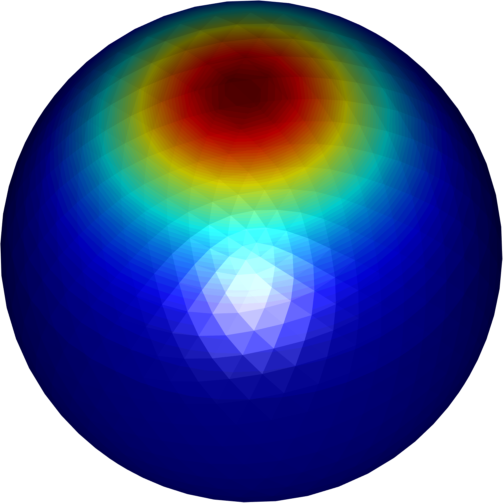}}}
\end{minipage}
\hspace{.2cm}
\begin{minipage}{0.6\textwidth}
\begin{tabular}{cccc}
\includegraphics[width=0.21\textwidth]{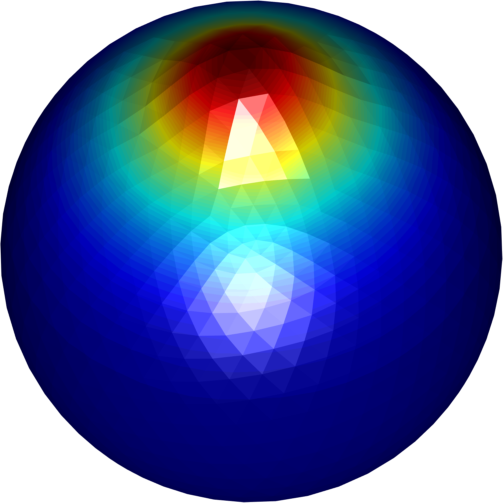}&
\includegraphics[width=0.21\textwidth]{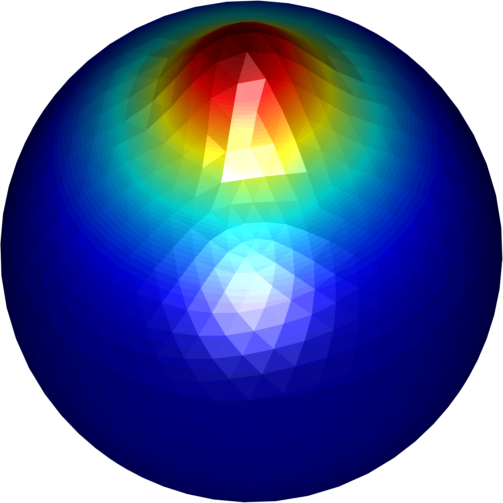}&
\includegraphics[width=0.21\textwidth]{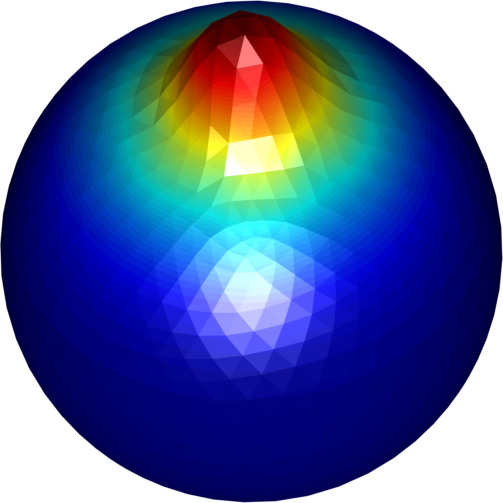}&
\includegraphics[width=0.21\textwidth]{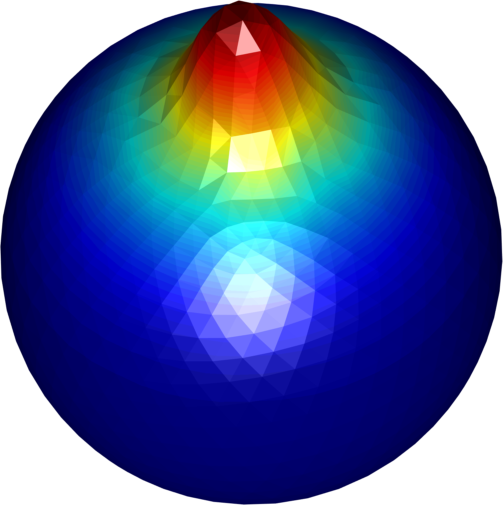} \\
\subfloat[$t=0.25$]{\includegraphics[width=0.21\textwidth]{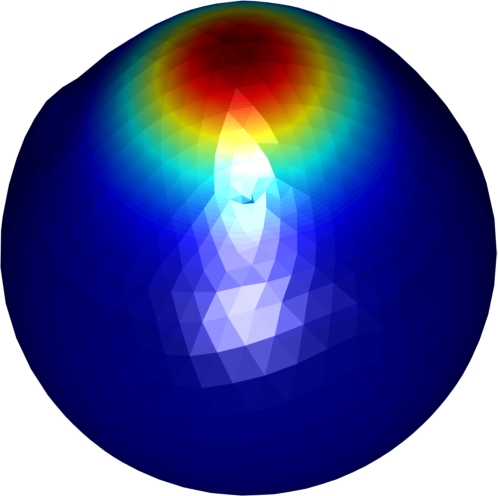}}&
\subfloat[$t=0.5$]{\includegraphics[width=0.21\textwidth]{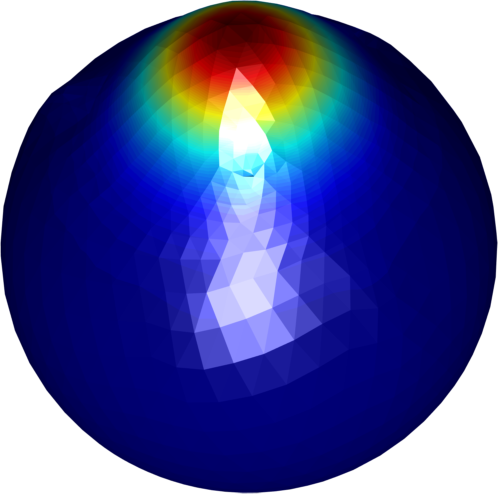}}&
\subfloat[$t=0.75$]{\includegraphics[width=0.21\textwidth]{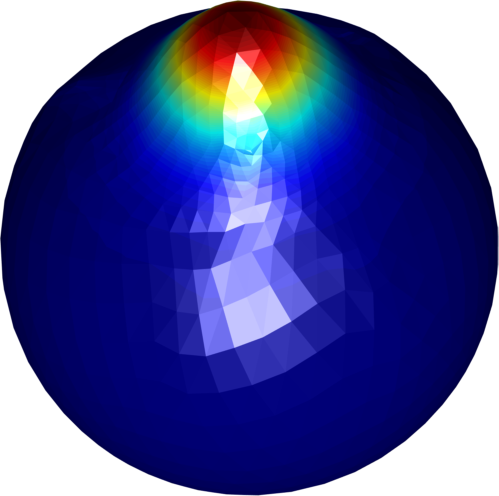}}&
\subfloat[$t=1$]{\includegraphics[width=0.21\textwidth]{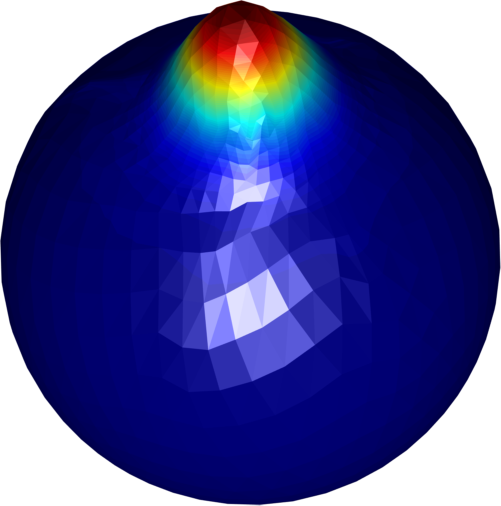}} 
\end{tabular}
\end{minipage}
\hspace{.2cm}
\begin{minipage}{0.13\textwidth}
\subfloat[Target $q_1$]{\fbox{\includegraphics[width=\textwidth]{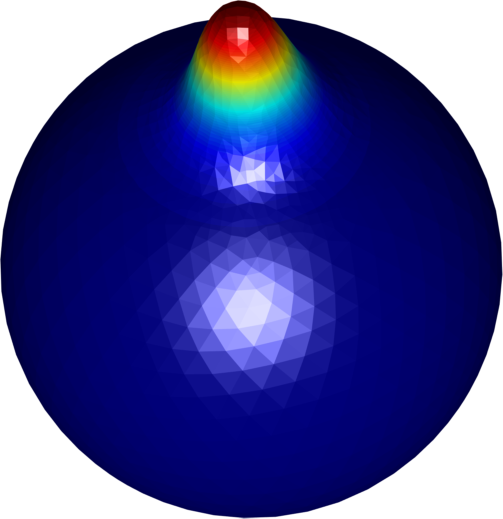}}}
\end{minipage}
\caption{Influence of texture on optimal point correspondences. Matching the textures requires compression of a large portion of a sphere $q_0$ (left) to a small portion of a deformed sphere $q_1$ (right). This compression occurs in the geodesic between the textured shapes (bottom) but not between the untextured shapes (top). It increases the SRNF distance between the matched surfaces $\tilde q_0$ and $\tilde q_1$ from 0.76 to 1.74, but this increase is compensated by a better fit of textures.}
\label{fig:fun_bump}       
\end{figure*}

\begin{figure*}
\centering\captionsetup[subfigure]{labelformat=empty}
\begin{minipage}{0.18\textwidth}
\subfloat[Source $q_0$]{\fbox{\includegraphics[width=\textwidth]{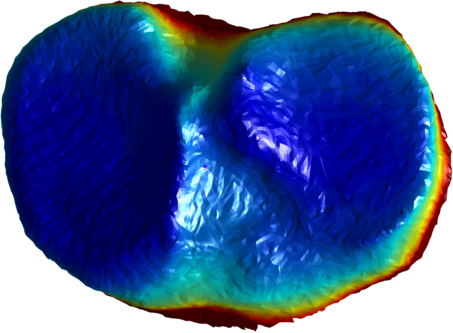}}}
\end{minipage}
\hspace{.1cm}
\begin{minipage}{0.6\textwidth}
\begin{tabular}{cccc}
\includegraphics[width=0.3\textwidth]{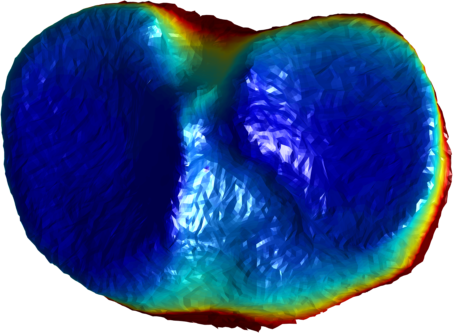}&
\includegraphics[width=0.3\textwidth]{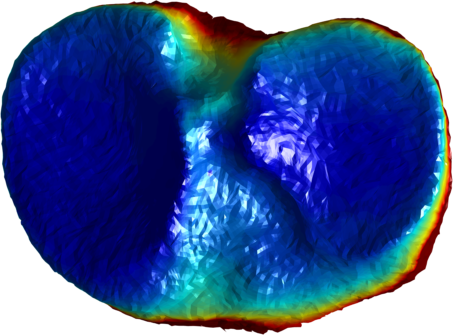}&
\includegraphics[width=0.3\textwidth]{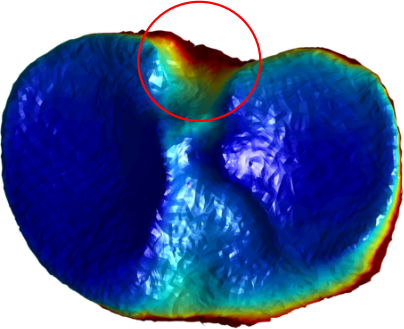} \\
\subfloat[$t=0.33$]{\includegraphics[width=0.3\textwidth]{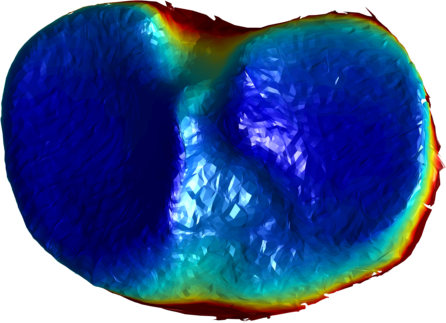}}&
\subfloat[$t=0.66$]{\includegraphics[width=0.3\textwidth]{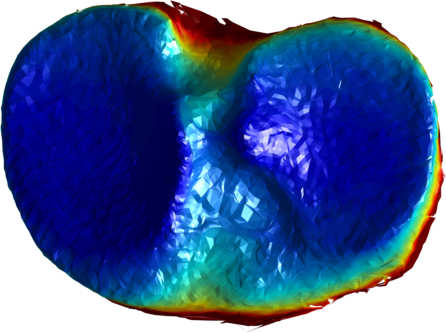}}&
\subfloat[$t=1$]{\includegraphics[width=0.3\textwidth]{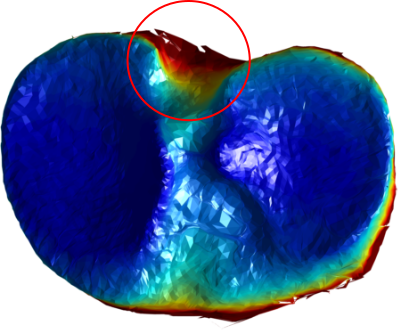}} 
\end{tabular}
\end{minipage}
\hspace{.1cm}
\begin{minipage}{0.18\textwidth}
\subfloat[Target $q_1$]{\fbox{\includegraphics[width=\textwidth]{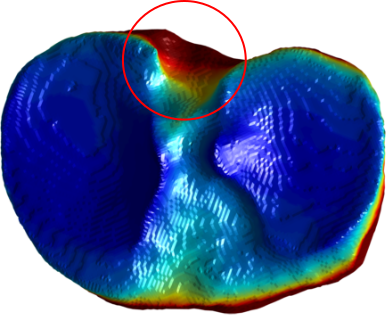}}}
\end{minipage}
\caption{Influence of texture on optimal point correspondences. Two tibia bones (left, right) carry texture information representing the width of the tibia-femoral joint. Differences in width are most pronounced in the upper part of the bone. This area is matched better by the geodesic interpolation of textured shapes (bottom) than of untextured shapes (top), see particular the area highlighted by the red circles.}
\label{fig:fun_knees}       
\end{figure*}

Our algorithm provides a convenient way of handling texture information, as explained in Section~\ref{sec:varifold}. 
Namely, the varifold distance can be augmented by an additional term which measures texture differences. 
Texture information can guide the optimizer in the matching problem and improve the resulting point correspondences. 
We demonstrate this in two examples, first on synthetic and then on real-world data.
 
 \begin{figure*}[h!]
\centering
\begin{tabular}{|c|c|c|}
\hline
\includegraphics[height=0.15\textheight]{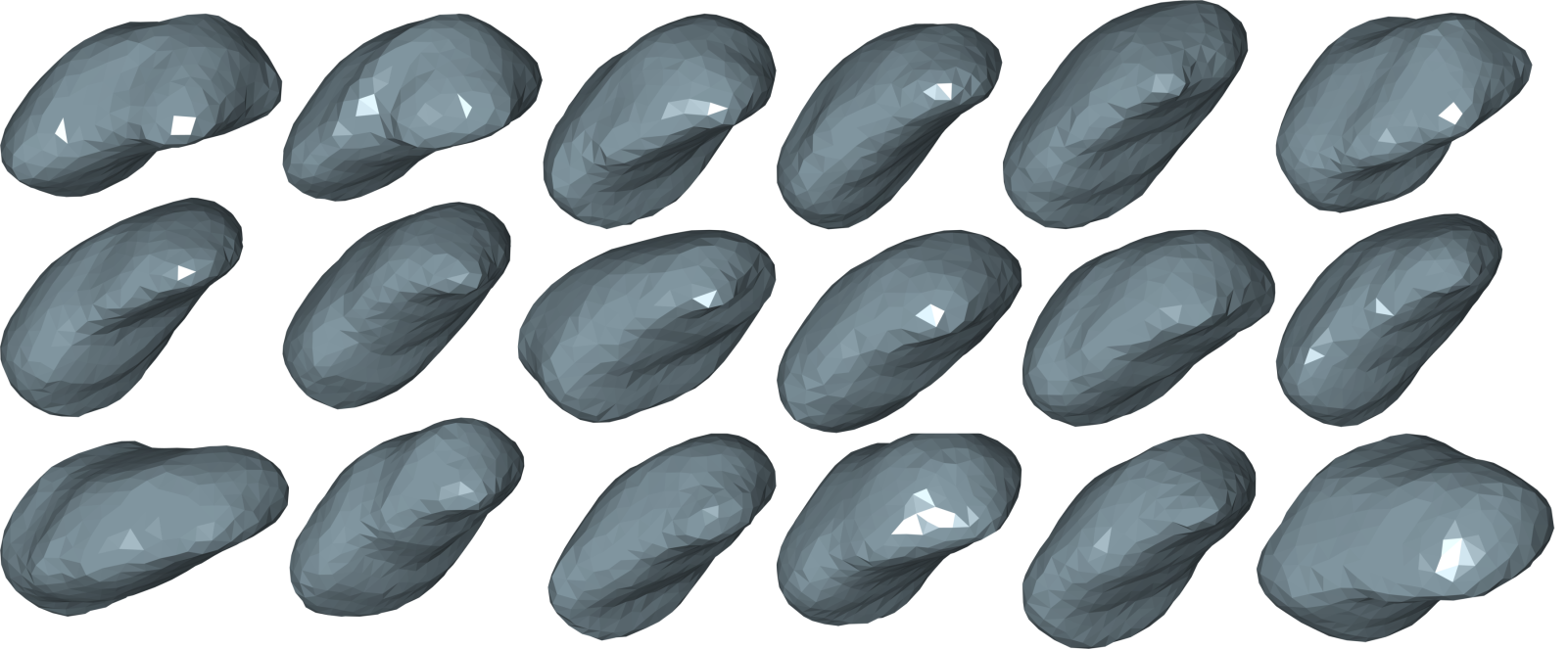}&
\includegraphics[height=0.15\textheight]{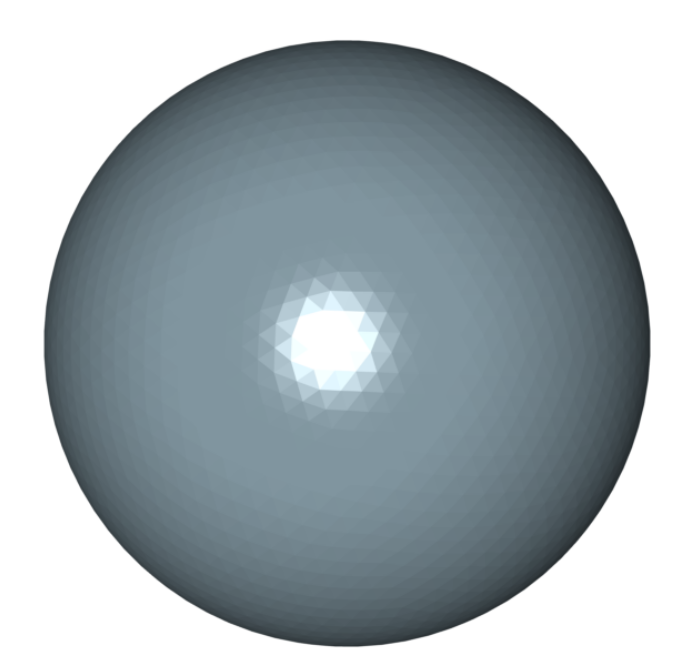}&
\includegraphics[height=0.15\textheight]{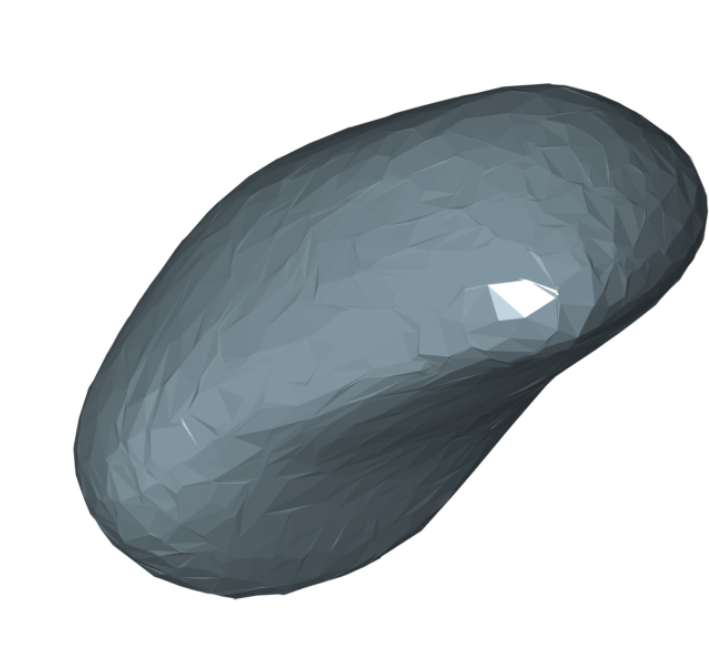}\\
Data& Initial Guess& Karcher\\
\hline
\end{tabular}
\caption{The Karcher mean (right) of a dataset of amygdala shapes (left), computed using Algorithm~\ref{alg:karcher} with a sphere (center) as initialization.}
\label{fig:frechet}
\end{figure*} 
As a synthetic example, we consider in Figure~\ref{fig:fun_bump} texture information which asks for compression of a large portion of a sphere to a small portion of a deformed sphere. 
The texture signals range from 0 to 1, and the texture kernel $\tau$ is Gaussian with a scale of $0.2$.
Without any texture information, there is no compression in the geodesic between the two shapes. 
If, however, texture information is included in the matching energy, then there is compression along the geodesic, in accordance with the texture information. 
Visually, the difference is most prominent in the terminal values of the two geodesics.
Quantitatively, the difference can also be seen from the resulting SRNF distance, which is higher when texture information is included (1.74 versus 0.76). 
The reason is that the optimizer has to strike a balance between matching the textures and minimizing the deformation energy.

The same qualitative behavior can be seen in the real-world example in Figure~\ref{fig:fun_knees}.
Here the data consist of tibia bone extremities of two subjects, and the texture represents the width of the tibia-femoral joint, estimated using the approach of \citet{Cao2015}. 
The texture signals range from 0 to 128, and the texture kernel is Gaussian with three successive scales of 100, 50, and 25. 
Differences in texture are most prominent in the upper middle part of the bone, where the red region is less pronounced in the left than in the right sample. 
These regions are matched incorrectly when no texture information is used (top), but are matched more accurately otherwise (bottom). 

\subsection{Population statistics}\label{sec:karcher}
Finally we show how our surface matching algorithm can be used to perform simple statistical analyses of shape populations, following the framework of geometric statistics; see for instance the recent review in~\cite{pennec2019riemannian}. 
A first example is multi-dimensional scaling or k-means clustering based on the pairwise distance matrix; see Figure~\ref{fig:cups_glasses}.
As a second example, we consider the estimation of the Karcher mean of a dataset of surface meshes. Figure~\ref{fig:frechet} shows the Karcher mean of 18 amygdala shapes from the BIOCARD data\-base \citep{BIOCARD2015} which were segmented from MRI images. To approximate the Karcher mean we followed a method adapted from~\cite{arnaudon2013approximating} and \cite{cury2013template}, which consists in an iterative geodesic centro\"{i}d procedure. The details of the algorithm are described in Algorithm~\ref{alg:karcher}. 

\begin{algorithm}
\Fn{\Karcher{$q_1$, \dots, $q_n$}}{
Initialize $\overline q$ to some mesh, which determines the topology/connectivity of the mean\;
\For{$i\longleftarrow 1$ \KwTo {\rm{some number of iterations}}}{
$(q_1, \dots q_n)\longleftarrow$\RandomShuffle{$q_1, \dots q_n$}\;
\For{$j\longleftarrow 1$ \KwTo $n$}{
$(\tilde q,\tilde q_j)\longleftarrow\MultiResolutionMatch(q,q_{j})$\;
$\overline N\longleftarrow N_{\tilde q}+\frac{1}{(i-1)*n+j}(N_{\tilde q_j}-N_{\tilde q})$\;
$\overline q\longleftarrow\InverseSRNF(\overline N)$
}
}
\KwRet{$\overline q$}
} 
\caption{Karcher mean}
\label{alg:karcher}
\end{algorithm}

\section{Discussion and conclusion}
\label{sec:conclusion}

The above examples demonstrate that our implementation is fast and versatile, and that it produces quite natural distances and interpolations between surfaces, including surfaces which have different topologies or carry texture information. These advantages are the result of our novel combination of elastic metrics and varifold distances. We also introduced the possibility of extending this idea to other surface feature maps, in particular square root curvature fields, as a way to emulate higher-order Sobolev metrics on surfaces. However, a precise study of the properties and effects of square root curvature fields in surface matching is left for future work.  

Besides these advantages, we want to mention some important caveats of our SRNF-based approach to surface matching. 
On the one hand, as the SRNF metric is only of first order and incomplete, 
minimizing the SRNF energy \eqref{equ:dist_srnf_shape} over reparametrizations can lead to shrinkage of some parts of a surface.
Accordingly, ``optimal reparametrizations'' may in fact be singular. 
On the other hand, varifold distances are by construction insensitive to structures of vanishing or small area, such as spikes or one-dimensional fibers. Thus, the combination of these two distances in our proposed formulation makes the optimizer in our matching procedure prone to finding 
surfaces with spiking singularities,
despite the SRNF distance still being computed accurately. 
These degeneracies are alleviated considerably by the multi-resolution minimization scheme which we introduced. Nevertheless, they can still be routinely observed in some simulations, most notably around irregular boundaries of objects.    

An important question for future work is whether this problem can be solved by suitable regularization, either using higher-order elastic metrics
(e.g.\ based on SRCFs) or higher-order fidelity metrics (e.g.\ normal cycles), which may not have the same vanishing behavior as varifolds. More generally, higher-order feature maps might also be of theoretical interest in order to overcome the incompleteness of low-order metrics.

\begin{acknowledgements}
The authors thank Stanley Durrleman, Jos\'{e} Braga, and Jean Dumoncel for the use of the cochlea data, Wojtek Zbijewski and his group for sharing the tibia surfaces as well as Daniel Tward and the BIOCARD team for sharing the amygdala dataset. In addition, we thank Zhe Su for his help with Figure~\ref{fig:shape_space} and the whole shape group at Florida State University for helpful discussions during the preparation of this manuscript. 
\end{acknowledgements}

\FloatBarrier

\appendix

\section{Notation}
\label{sec:notation}

Throughout this section, $h \in T_q\Imm=C^\infty(M,\mathbb R^3)$ denotes a tangent vector to $q\in\Imm$, and $X,Y$ are vector fields on $M$. 
Traces are denoted by $\Tr$ or a dot, $T$ is the tangent functor, and $D$ stands for directional derivatives.
For instance, the derivative at $q$ in the direction of $h$ is denoted by $D_{(q,h)}$. 
We write $\overline g=\langle\cdot,\cdot\rangle$ for the Euclidean inner product on $\mathbb R^3$, 
$|\cdot|$ for the Euclidean norm on $\mathbb R^3$,
$g_q=q^*\overline g$ for the pull-back metric on $TM$, 
$g_q^{-1}$ for the cometric on $T^*M$, 
and $g_q^{-1}\otimes\overline g$ for the product metric on $T^*M\otimes\mathbb R^3$.
The metric $g_q$ corresponds to a fiber-linear map $\flat$ from $TM$ to $T^*M$, and the cometric $g_q^{-1}$ corresponds to a fiber-linear map $\sharp$ from $T^*M$ to $M$.
The Riemannian surface measure of $g_q$ is denoted by $\vol_q$, and the corresponding half-density by $\vol_q^{1/2}$.
The surface measure is a section of the volume bundle $\Vol$, and the half-density of the half-density bundle $\Vol^{1/2}$.
The normal projection $\bot\colon M\to L(\mathbb R^3,\mathbb R^3)$ is defined as $\bot=\langle\cdot,n_q\rangle n_q$, 
the tangential projection $\top\colon M\to L(\mathbb R^3,TM)$ is defined as $\top=(Tq)^{-1}(\operatorname{Id}_{\mathbb R^3}-\bot)$, and one has the identity $\bot+Tq\circ\top=\Id_{\mathbb R^3}$. 
Depending on the context, $\nabla$ is the covariant derivative on $\mathbb R^3$, which coincides with the usual coordinate derivative, or the Levi-Civita covariant derivative of $g_q$.
For instance, in the definition $\nabla^2_{X,Y}h\coloneqq \nabla_X\nabla_Yh-\nabla_{\nabla_XY}h$, only $\nabla_XY$ is the Levi-Civita covariant derivative, and all other derivatives are coordinate derivatives. 

\section{Formula for SRNF metrics}
\label{sec:srnf_formula}

In this section we establish the explicit formula \eqref{equ:srnf_metric_formula} for the SRNF metric. 
We need some variational formulas from e.g.\  \citet{bauer2012almost}:
\begin{align*}
D_{(q,h)}\vol_q 
=&
\Tr\big(g_q^{-1}.\langle\nabla h,Tq\rangle\big)\vol_q
=
\Tr\big((\nabla h)^\top\big)\;,
\\
D_{(q,h)}\vol_q^{1/2}
=&
\tfrac12\Tr\big((\nabla h)^\top\big)\vol_q^{1/2}\;,
\\
D_{(q,h)}n_q
=&
-Tq\circ\langle n_q,\nabla h\rangle^\sharp\;, 
\\
D_{(q,h)}\srnf_q
=&
\Big(-Tq\circ\langle n_q,\nabla h\rangle^\sharp
+n_q \tfrac12\Tr\big((\nabla h)^\top\big)\Big)\vol_q^{1/2}\;.
\end{align*}
Putting these together, one obtains the following expression for the SRNF metric \eqref{equ:srnf_metric}:
\begin{align*}
G_q(h,h)
\coloneqq &
\int_M |D_{(q,h)}\srnf_q|^2
\\=&
\int_M|Tq\circ\langle n_q,\nabla h\rangle^\sharp|^2\vol_q
\\&+
\int_M|n_q \tfrac12\Tr\big((\nabla h)^\top\big)|^2\vol_q
\\=&
\int_M|\langle n_q,\nabla h\rangle|_{g_q^{-1}}^2\vol_q
+
\frac14\int_M\Tr\big((\nabla h)^\top\big)^2\vol_q
\\=&
\int_M|(\nabla h)^\bot|_{g_q^{-1}\otimes\overline{g}}^2\vol_q
+
\frac14\int_M\Tr\big((\nabla h)^\top\big)^2\vol_q\;.
\end{align*}

\section{Approximation properties of SRNF distances}
\label{sec:approximation}

This section makes precise in what sense the SRNF distance approximates the geodesic distance of the SRNF metric.
The geodesic distance of the SRNF metric is lower-bounded by the SRNF distance because the latter is a chordal distance:
\begin{align*}
\dist_{\Imm}(q_0,q_1)^2
\coloneqq&
\inf_q \int_0^1 G_q(\partial_t q,\partial_t q) dt
\\=&
\inf_q \int_0^1 |D_{(q,\partial_t q)}\srnf_q|^2 dt
\geq 
\|\srnf_{q_1}-\srnf_{q_0}\|_{L^2}^2\;,
\end{align*}
where the infimum is over all paths $q$ as in \eqref{equ:dist_imm}. 
Conversely, the geodesic distance of the SRNF metric is upper-bounded by the length of the linear interpolation between the immersions $q_0$ and $q_1$, provided they are sufficiently close to each other for this to make sense, leading to the upper bound 
\begin{align*}
\dist_{\Imm}(q_0,q_1)^2
&\leq
G_{q_0}(q_1-q_0,q_1-q_0)
\\&=
\|D_{(q_0,q_1-q_0)}\srnf_{q_0}\|_{L^2}^2
\approx
\|\srnf_{q_1}-\srnf_{q_0}\|_{L^2}^2\;,
\end{align*}
which is valid in any chart around $q_0$ up to terms of order $o(\|\srnf_{q_1}-\srnf_{q_0}\|_{L^2}^2)$. 

\section{Formula for SRCF metrics}
\label{sec:srcf_metric}

Recall that the scalar and vector-valued second fundamental forms are defined as 
\begin{align*}
s_q(X,Y)
&\coloneqq
\langle \nabla_X(Tq\circ Y),n_q\rangle\;, 
&
S_q(X,Y)&\coloneqq s_q(X,Y)n_q\;,
\end{align*}
for any vector fields $X$ and $Y$ on $M$. 
Following \citet{bauer2012almost}, one obtains the variational formulas
\begin{align*}
\hspace{2em}&\hspace{-2em}
D_{(q,h)} s_q(X,Y)
=
\langle D_{(q,h)}\nabla_X(Tq\circ Y),n_q\rangle
\\&\qquad
+\langle \nabla_X(Tq\circ Y),D_{(q,h)}n_q\rangle
\\&=
\langle \nabla_X\nabla_Yh,n_q\rangle
-\langle \nabla_X(Tq\circ Y),Tq\circ\langle\nabla h,n_q\rangle^\sharp\rangle
\\&=
\langle \nabla_X\nabla_Yh,n_q\rangle
-g_q( \nabla_XY,\langle\nabla h,n_q\rangle^\sharp)
\\&=
\langle \nabla_X\nabla_Yh,n_q\rangle
-\langle\nabla_{\nabla_XY} h,n_q\rangle
=
\langle \nabla^2_{X,Y}h,n_q\rangle\;,
\\
\hspace{2em}&\hspace{-2em}
D_{(q,h)} S_q(X,Y)
=
(D_{(q,h)} s_q(X,Y))n_q
+
s_q(X,Y)(D_{(q,h)} n_q)
\\&=
(\nabla^2_{X,Y}h)^\bot
-s_q(X,Y)Tq\circ\langle\nabla h,n_q\rangle^\sharp\;.
\end{align*} 
Using the formula $\Delta_q=-\Tr(g_q^{-1}\nabla^2)$ for the Laplacian, this yields the following variational formula for the vector-valued mean curvature $H_q=\Tr(g_q^{-1}S_q)$: 
\begin{align*}
D_{(q,h)}g_q
&=
D_{(q,h)}\langle Tq,Tq\rangle
=
\langle \nabla h,Tq\rangle
+
\langle Tq,\nabla h\rangle\;,
\\
D_{(q,h)}g_q^{-1}
&=
-g_q^{-1}(D_{(q,h)}g_q)g_q^{-1}\;,
\\
D_{(q,h)}H_q
&=
\Tr(g_q^{-1}D_{(q,h)}S_q)
+
\Tr(S_qD_{(q,h)}g_q^{-1})
\\&=
-(\Delta_q h)^\bot
-\Tr(g_q^{-1}s_q)Tq\circ\langle\nabla h,n_q\rangle^\sharp
\\&\qquad
-2\Tr(S_qg_q^{-1}\langle \nabla h,Tq\rangle g_q^{-1})\;.
\end{align*}
Letting $C(\nabla h)$ denote the first-order terms in $-D_{(q,h)}H_q$, one obtains the desired formula for the SRCF metric:
\begin{align*}
G_q(h,h)
&\coloneqq
\int_M|D_{(q,h)}H_q|^2\vol_q
=
\int_M |(\Delta_qh)^\bot+C(\nabla h)|^2\vol_q\;.
\end{align*}

\bibliographystyle{spbasic_updated}

\begin{thebibliography}{76}
\providecommand{\natexlab}[1]{#1}
\providecommand{\url}[1]{{#1}}
\providecommand{\urlprefix}{URL }
\expandafter\ifx\csname urlstyle\endcsname\relax
  \providecommand{\doi}[1]{DOI~\discretionary{}{}{}#1}\else
  \providecommand{\doi}{DOI~\discretionary{}{}{}\begingroup
  \urlstyle{rm}\Url}\fi
\providecommand{\eprint}[2][]{\url{#2}}

\bibitem[{Abe and Erbacher(1975)}]{abe1975isometric}
Abe K, Erbacher J (1975) Isometric immersions with the same gauss map.
  Mathematische Annalen 215(3):197--201

\bibitem[{Almgren(1966)}]{Almgren1966}
Almgren F (1966) Plateau's Problem: An Invitation to Varifold Geometry. Student
  Mathematical Library

\bibitem[{Arnaudon and Nielsen(2013)}]{arnaudon2013approximating}
Arnaudon M, Nielsen F (2013) {On approximating the Riemannian 1-center}.
  Computational Geometry 46(1):93--104

\bibitem[{Bauer et~al.(2011)Bauer, Harms, and Michor}]{bauer2011sobolev}
Bauer M, Harms P, Michor PW (2011) Sobolev metrics on shape space of surfaces.
  J Geom Mech 3(4):389--438

\bibitem[{Bauer et~al.(2012)Bauer, Harms, and Michor}]{bauer2012almost}
Bauer M, Harms P, Michor PW (2012) Almost local metrics on shape space of
  hypersurfaces in n-space. SIAM J Imaging Sci 5(1):244--310

\bibitem[{Bauer et~al.(2014)Bauer, Bruveris, and Michor}]{bauer2014overview}
Bauer M, Bruveris M, Michor PW (2014) Overview of the geometries of shape
  spaces and diffeomorphism groups. J Math Imaging Vision 50(1-2):60--97

\bibitem[{Bauer et~al.(2016)Bauer, Bruveris, and Michor}]{bauer2016use}
Bauer M, Bruveris M, Michor PW (2016) Why use {Sobolev} metrics on the space of
  curves. In: Riemannian Computing in Computer Vision, Springer, pp 233--255

\bibitem[{Bauer et~al.(2017)Bauer, Bruveris, Charon, and
  M{\o}ller-Andersen}]{bauer2017varifold}
Bauer M, Bruveris M, Charon N, M{\o}ller-Andersen J (2017) Varifold-based
  matching of curves via sobolev-type riemannian metrics. In: Graphs in
  Biomedical Image Analysis, Computational Anatomy and Imaging Genetics,
  Springer, pp 152--163

\bibitem[{Bauer et~al.(2019{\natexlab{a}})Bauer, Bruveris, Charon, and
  M{\o}ller-Andersen}]{bauer2018relaxed}
Bauer M, Bruveris M, Charon N, M{\o}ller-Andersen J (2019{\natexlab{a}}) A
  relaxed approach for curve matching with elastic metrics. ESAIM: Control,
  Optimisation and Calculus of Variations 25:72

\bibitem[{Bauer et~al.(2019{\natexlab{b}})Bauer, Charon, and
  Harms}]{bauer2019inexact}
Bauer M, Charon N, Harms P (2019{\natexlab{b}}) Inexact elastic shape matching
  in the square root normal field framework. In: International Conference on
  Geometric Science of Information, Springer, pp 13--20

\bibitem[{Bauer et~al.(2020)Bauer, Harms, and Michor}]{bauer2020fractional}
Bauer M, Harms P, Michor PW (2020) Fractional {Sobolev} metrics on spaces of
  immersions. Calculus of Variations and Partial Differential Equations 59(62)

\bibitem[{Beg et~al.(2005)Beg, Miller, Trouv{\'e}, and
  Younes}]{beg2005computing}
Beg MF, Miller MI, Trouv{\'e} A, Younes L (2005) Computing large deformation
  metric mappings via geodesic flows of diffeomorphisms. International journal
  of computer vision 61(2):139--157

\bibitem[{Bernal et~al.(2016)Bernal, Dogan, and Hagwood}]{bernal2016fast}
Bernal J, Dogan G, Hagwood CR (2016) Fast dynamic programming for elastic
  registration of curves. In: Computer Vision and Pattern Recognition (CVPR),
  pp 111--118

\bibitem[{Bhattacharya and Bhattacharya(2012)}]{bhattacharya2012nonparametric}
Bhattacharya A, Bhattacharya R (2012) Nonparametric inference on manifolds:
  with applications to shape spaces, vol~2. Cambridge University Press

\bibitem[{Bobenko et~al.(2008)Bobenko, Schr{\"o}der, Sullivan, and
  Ziegler}]{bobenko2008discrete}
Bobenko A, Schr{\"o}der P, Sullivan JM, Ziegler GM (eds)  (2008) Discrete
  differential geometry. No.~38 in {Oberwolfach Semin.}, Birkh{\"a}user

\bibitem[{Bruveris(2015)}]{bruveris2015completeness}
Bruveris M (2015) Completeness properties of {Sobolev} metrics on the space of
  curves. Journal of Geometric Mechanics 7(2):125--150

\bibitem[{Bruveris et~al.(2014)Bruveris, Michor, and
  Mumford}]{bruveris2014geodesic}
Bruveris M, Michor PW, Mumford D (2014) Geodesic completeness for {Sobolev}
  metrics on the space of immersed plane curves. In: Forum of Mathematics,
  Sigma, Cambridge University Press, vol~2

\bibitem[{Cao et~al.(2015)Cao, Thawait, Gang, Zbijewski, Reigel, Brown, Corner,
  Demehri, and Siewerdsen}]{Cao2015}
Cao Q, Thawait G, Gang GJ, Zbijewski W, Reigel T, Brown T, Corner B, Demehri S,
  Siewerdsen JH (2015) {Characterization of 3D joint space morphology using an
  electrostatic model (with application to osteoarthritis).} Physics in
  medicine and biology 60(3):947--60

\bibitem[{Cervera et~al.(1991)Cervera, Mascar{\'o}, and
  Michor}]{cervera1991action}
Cervera V, Mascar{\'o} F, Michor PW (1991) {The action of the diffeomorphism
  group on the space of immersions}. Differential Geom Appl 1(4):391--401

\bibitem[{Charlier et~al.(2017)Charlier, Charon, and
  Trouv{\'e}}]{charlier2017fshape}
Charlier B, Charon N, Trouv{\'e} A (2017) The fshape framework for the
  variability analysis of functional shapes. Found Comput Math 17(2):287--357

\bibitem[{Charlier et~al.(2020)Charlier, Feydy, Glaun{\`e}s, Collin, and
  Durif}]{charlier2020kernel}
Charlier B, Feydy J, Glaun{\`e}s J, Collin FD, Durif G (2020) Kernel operations
  on the {GPU}, with autodiff, without memory overflows. arXiv preprint
  arXiv:200411127

\bibitem[{Charon and Trouv{\'e}(2013)}]{charon2013varifold}
Charon N, Trouv{\'e} A (2013) The varifold representation of nonoriented shapes
  for diffeomorphic registration. SIAM J Imaging Sci 6(4):2547--2580

\bibitem[{Charon and Trouv{\'e}(2014)}]{charon2014functional}
Charon N, Trouv{\'e} A (2014) Functional currents: a new mathematical tool to
  model and analyse functional shapes. J Math Imaging Vision 48(3):413--431

\bibitem[{Charon et~al.(2020)Charon, Charlier, Glaun\`{e}s, Gori, and
  Roussillon}]{Charon2020fidelity}
Charon N, Charlier B, Glaun\`{e}s J, Gori P, Roussillon P (2020) Fidelity
  metrics between curves and surfaces: currents, varifolds, and normal cycles.
  In: Riemannian Geometric Statistics in Medical Image Analysis, Academic
  Press, pp 441 -- 477

\bibitem[{Cury et~al.(2013)Cury, Glaunes, and Colliot}]{cury2013template}
Cury C, Glaunes JA, Colliot O (2013) Template estimation for large database: a
  diffeomorphic iterative centroid method using currents. In: International
  Conference on Geometric Science of Information, Springer, pp 103--111

\bibitem[{Dryden and Mardia(1998)}]{dryden1998statistical}
Dryden IL, Mardia KV (1998) {Statistical shape analysis}. {Wiley Series in
  Probability and Statistics: Probability and Statistics}, John Wiley \& Sons
  Ltd., Chichester

\bibitem[{Ebin(1970)}]{Ebin1970b}
Ebin DG (1970) The manifold of {R}iemannian metrics. In: Global {A}nalysis
  ({P}roc. {S}ympos. {P}ure {M}ath., {V}ol. {XV}, {B}erkeley, {C}alif., 1968),
  Amer. Math. Soc., Providence, R.I., pp 11--40

\bibitem[{Federer(1969)}]{federer1969geometric}
Federer H (1969) Geometric measure theory. Springer

\bibitem[{Floater and Hormann(2005)}]{floater2005surface}
Floater MS, Hormann K (2005) Surface parameterization: a tutorial and survey.
  In: Advances in multiresolution for geometric modelling, Springer, pp
  157--186

\bibitem[{Frenkel and Basri(2003)}]{frenkel2003curve}
Frenkel M, Basri R (2003) Curve matching using the fast marching method. In:
  International Workshop on Energy Minimization Methods in Computer Vision and
  Pattern Recognition, Springer, pp 35--51

\bibitem[{Fr{\"o}hlich and Botsch(2011)}]{frohlich2011example}
Fr{\"o}hlich S, Botsch M (2011) Example-driven deformations based on discrete
  shells. In: Computer graphics forum, Wiley Online Library, vol~30, pp
  2246--2257

\bibitem[{Geirhos et~al.(2018)Geirhos, Rubisch, Michaelis, Bethge, Wichmann,
  and Brendel}]{geirhos2018imagenet}
Geirhos R, Rubisch P, Michaelis C, Bethge M, Wichmann FA, Brendel W (2018)
  {ImageNet-trained CNNs are biased towards texture; increasing shape bias
  improves accuracy and robustness}. arXiv preprint arXiv:181112231

\bibitem[{Glaun{\`e}s et~al.(2008)Glaun{\`e}s, Qiu, Miller, and
  Younes}]{glaunes2008large}
Glaun{\`e}s J, Qiu A, Miller MI, Younes L (2008) Large deformation
  diffeomorphic metric curve mapping. Int J Comput Vis 80(3):317

\bibitem[{Grzegorzek et~al.(2013)Grzegorzek, Theobalt, Koch, and
  Kolb}]{grzegorzek2013time}
Grzegorzek M, Theobalt C, Koch R, Kolb A (2013) Time-of-Flight and Depth
  Imaging. Sensors, Algorithms and Applications: Dagstuhl Seminar 2012 and GCPR
  Workshop on Imaging New Modalities, vol 8200. Springer

\bibitem[{Jermyn et~al.(2012)Jermyn, Kurtek, Klassen, and
  Srivastava}]{jermyn2012elastic}
Jermyn IH, Kurtek S, Klassen E, Srivastava A (2012) Elastic shape matching of
  parameterized surfaces using square root normal fields. In: European
  Conference on Computer Vision, Springer, pp 804--817

\bibitem[{Jermyn et~al.(2017)Jermyn, Kurtek, Laga, and
  Srivastava}]{jermyn2017elastic}
Jermyn IH, Kurtek S, Laga H, Srivastava A (2017) Elastic shape analysis of
  three-dimensional objects. Synthesis Lectures on Computer Vision 12(1):1--185

\bibitem[{Kaltenmark et~al.(2017)Kaltenmark, Charlier, and
  Charon}]{kaltenmark2017general}
Kaltenmark I, Charlier B, Charon N (2017) A general framework for curve and
  surface comparison and registration with oriented varifolds. In: Computer
  Vision and Pattern Recognition (CVPR)

\bibitem[{Kendall et~al.(1999)Kendall, Barden, Carne, and Le}]{Kendall1999}
Kendall DG, Barden D, Carne TK, Le H (1999) {Shape and shape theory}. {Wiley
  Series in Probability and Statistics}, John Wiley \& Sons Ltd., Chichester,
  \doi{10.1002/9780470317006}

\bibitem[{Kilian et~al.(2007)Kilian, Mitra, and Pottmann}]{kilian2007geometric}
Kilian M, Mitra NJ, Pottmann H (2007) Geometric modeling in shape space. ACM
  Trans Graphics 26(3)

\bibitem[{Klassen and Michor(2020)}]{klassen2019closed}
Klassen E, Michor PW (2020) Closed surfaces with different shapes that are
  indistinguishable by the srnf. Archivum Mathematicum 56(2):107--114

\bibitem[{Kokoszka and Reimherr(2017)}]{kokoszka2017introduction}
Kokoszka P, Reimherr M (2017) Introduction to functional data analysis. CRC
  Press

\bibitem[{Kurtek et~al.(2012)Kurtek, Klassen, Gore, Ding, and
  Srivastava}]{kurtek2012elastic}
Kurtek S, Klassen E, Gore JC, Ding Z, Srivastava A (2012) Elastic geodesic
  paths in shape space of parameterized surfaces. IEEE Trans Pattern Anal Mach
  Intell 34(9):1717--1730

\bibitem[{Laga et~al.(2017)Laga, Xie, Jermyn, and
  Srivastava}]{laga2017numerical}
Laga H, Xie Q, Jermyn IH, Srivastava A (2017) {Numerical inversion of SRNF maps
  for elastic shape analysis of genus-zero surfaces}. IEEE Trans Pattern Anal
  Mach Intell 39(12):2451--2464

\bibitem[{Lahiri et~al.(2015)Lahiri, Robinson, and Klassen}]{lahiri2015precise}
Lahiri S, Robinson D, Klassen E (2015) Precise matching of {PL} curves in
  {$\mathbb R^N$} in the square root velocity framework. Geometry, Imaging and
  Computing 2(3):133--186

\bibitem[{Liu and Nocedal(1989)}]{liu1989limited}
Liu DC, Nocedal J (1989) On the limited memory bfgs method for large scale
  optimization. Mathematical programming 45(1-3):503--528

\bibitem[{Marin et~al.(2019)Marin, Melzi, Rodol{\`a}, and
  Castellani}]{marin2019high}
Marin R, Melzi S, Rodol{\`a} E, Castellani U (2019) High-resolution
  augmentation for automatic template-based matching of human models. In: 2019
  International Conference on 3D Vision (3DV), IEEE, pp 230--239

\bibitem[{Mennucci et~al.(2008)Mennucci, Yezzi, and
  Sundaramoorthi}]{mennucci2008properties}
Mennucci A, Yezzi A, Sundaramoorthi G (2008) Properties of {Sobolev}-type
  metrics in the space of curves. Interfaces Free Bound 10(4):423--445

\bibitem[{Michor(2008)}]{Michor08}
Michor PW (2008) Topics in differential geometry, Graduate Studies in
  Mathematics, vol~93. American Mathematical Society, Providence, RI

\bibitem[{Michor and Mumford(2005)}]{michor2005vanishing}
Michor PW, Mumford D (2005) Vanishing geodesic distance on spaces of
  submanifolds and diffeomorphisms. Doc Math 10:217--245

\bibitem[{Michor and Mumford(2006)}]{michor2006riemannian}
Michor PW, Mumford D (2006) Riemannian geometries on spaces of plane curves.
  Journal of the European Mathematical Society 8:1--48

\bibitem[{Michor and Mumford(2007)}]{michor2007overview}
Michor PW, Mumford D (2007) An overview of the {Riemannian} metrics on spaces
  of curves using the {Hamiltonian} approach. Appl Comput Harmon Anal
  23(1):74--113

\bibitem[{Miller et~al.(2015)Miller, Ratnanather, Tward, Brown, Lee, Ketcha,
  Mori, Wang, Mori, Albert, Younes, and ~}]{BIOCARD2015}
Miller M, Ratnanather JT, Tward DJ, Brown T, Lee D, Ketcha M, Mori K, Wang MC,
  Mori S, Albert M, Younes L, ~ BRT (2015) {Network Neurodegeneration in
  Alzheimer’s Disease via MRI Based Shape Diffeomorphometry and High-Field
  Atlasing}. Frontiers in Bioengineering and Biotechnology 3:54

\bibitem[{Minh et~al.(2016)Minh, Murino, and Minh}]{minh2016algorithmic}
Minh HQ, Murino V, Minh HQ (2016) Algorithmic Advances in Riemannian Geometry
  and Applications. Springer

\bibitem[{Needham and Kurtek(2020)}]{needham2020simplifying}
Needham T, Kurtek S (2020) {Simplifying Transforms for General Elastic Metrics
  on the Space of Plane Curves}. SIAM Journal on Imaging Sciences
  13(1):445--473

\bibitem[{Niethammer et~al.(2019)Niethammer, Kwitt, and
  Vialard}]{niethammer2019metric}
Niethammer M, Kwitt R, Vialard FX (2019) Metric learning for image
  registration. In: Proceedings of the IEEE Conference on Computer Vision and
  Pattern Recognition, pp 8463--8472

\bibitem[{Pennec et~al.(2019)Pennec, Sommer, and
  Fletcher}]{pennec2019riemannian}
Pennec X, Sommer S, Fletcher T (2019) Riemannian Geometric Statistics in
  Medical Image Analysis. Academic Press

\bibitem[{Roussillon and Glaunes(2016)}]{roussillon2016kernel}
Roussillon P, Glaunes JA (2016) Kernel metrics on normal cycles and application
  to curve matching. SIAM J Imaging Sci 9(4):1991--2038

\bibitem[{Rumpf and Wardetzky(2014)}]{rumpf2014geometry}
Rumpf M, Wardetzky M (2014) Geometry processing from an elastic perspective.
  GAMM-Mitteilungen 37(2):184--216

\bibitem[{Rumpf and Wirth(2015{\natexlab{a}})}]{rumpf2015bvariational}
Rumpf M, Wirth B (2015{\natexlab{a}}) Variational methods in shape analysis.
  Handbook of Mathematical Methods in Imaging 2:1819--1858

\bibitem[{Rumpf and Wirth(2015{\natexlab{b}})}]{rumpf2015variational}
Rumpf M, Wirth B (2015{\natexlab{b}}) Variational time discretization of
  geodesic calculus. IMA Journal of Numerical Analysis 35(3):1011--1046

\bibitem[{Sebastian et~al.(2003)Sebastian, Klein, and
  Kimia}]{sebastian2003aligning}
Sebastian TB, Klein PN, Kimia BB (2003) On aligning curves. IEEE transactions
  on pattern analysis and machine intelligence 25(1):116--125

\bibitem[{Sheffer et~al.(2007)Sheffer, Praun, Rose et~al.}]{sheffer2007mesh}
Sheffer A, Praun E, Rose K, et~al. (2007) Mesh parameterization methods and
  their applications. Foundations and Trends in Computer Graphics and Vision
  2(2):105--171

\bibitem[{Srivastava and Klassen(2016)}]{srivastava2016functional}
Srivastava A, Klassen EP (2016) Functional and shape data analysis. Springer

\bibitem[{Srivastava et~al.(2011)Srivastava, Klassen, Joshi, and
  Jermyn}]{srivastava2011shape}
Srivastava A, Klassen E, Joshi SH, Jermyn IH (2011) Shape analysis of elastic
  curves in {Euclidean} spaces. IEEE Trans Pattern Anal Mach Intell
  33(7):1415--1428

\bibitem[{Su et~al.(2020{\natexlab{a}})Su, Bauer, Gallivan, and
  Klassen}]{su2020simplifying}
Su Z, Bauer M, Gallivan KA, Klassen E (2020{\natexlab{a}}) Simplifying
  transformations for a family of elastic metrics on the space of surfaces. In:
  IEEE Conference on Computer Vision and Pattern Recognition Workshops (CVPRW)

\bibitem[{Su et~al.(2020{\natexlab{b}})Su, Bauer, Preston, Laga, and
  Klassen}]{su2020shape}
Su Z, Bauer M, Preston SC, Laga H, Klassen E (2020{\natexlab{b}}) Shape
  analysis of surfaces using general elastic metrics. Journal of Mathematical
  Imaging and Vision 62:1087--1106

\bibitem[{Sullivan(2008)}]{sullivan2008curvatures}
Sullivan JM (2008) Curvatures of smooth and discrete surfaces. In:
  \cite{bobenko2008discrete}, pp 175--188

\bibitem[{Sundaramoorthi et~al.(2011)Sundaramoorthi, Mennucci, Soatto, and
  Yezzi}]{sundaramoorthi2011new}
Sundaramoorthi G, Mennucci A, Soatto S, Yezzi A (2011) A new geometric metric
  in the space of curves, and applications to tracking deforming objects by
  prediction and filtering. SIAM J Imaging Sci 4(1):109--145

\bibitem[{Tumpach(2016)}]{tumpach2016gauge}
Tumpach AB (2016) Gauge invariance of degenerate riemannian metrics. Notices of
  the AMS 63(4)

\bibitem[{Tumpach et~al.(2015)Tumpach, Drira, Daoudi, and
  Srivastava}]{tumpach2015gauge}
Tumpach AB, Drira H, Daoudi M, Srivastava A (2015) Gauge invariant framework
  for shape analysis of surfaces. IEEE transactions on pattern analysis and
  machine intelligence 38(1):46--59

\bibitem[{Turaga and Srivastava(2016)}]{turaga2016riemannian}
Turaga PK, Srivastava A (2016) Riemannian computing in computer vision.
  Springer

\bibitem[{Vaillant and Glaun{\`e}s(2005)}]{vaillant2005surface}
Vaillant M, Glaun{\`e}s J (2005) Surface matching via currents. In: Biennial
  International Conference on Information Processing in Medical Imaging,
  Springer, pp 381--392

\bibitem[{Willmore(1993)}]{willmore1993riemannian}
Willmore TJ (1993) Riemannian geometry. Oxford University Press

\bibitem[{Younes(1998)}]{younes1998computable}
Younes L (1998) Computable elastic distances between shapes. SIAM J Appl Math
  58(2):565--586

\bibitem[{Younes(2010)}]{younes2010shapes}
Younes L (2010) Shapes and diffeomorphisms, vol 171. Springer Science \&
  Business Media

\bibitem[{Younes et~al.(2008)Younes, Michor, Shah, and
  Mumford}]{younes2008metric}
Younes L, Michor PW, Shah J, Mumford D (2008) A metric on shape space with
  explicit geodesics. Atti Accad Naz Lincei Rend Lincei Mat Appl 19(1):25--57

\end{thebibliography}

\end{document}